\documentclass[10pt,journal,compsoc]{IEEEtran}
\usepackage{graphicx}
\usepackage{subfigure}
\usepackage{makecell}
\usepackage{amsmath,amssymb,amsfonts,bm}
\usepackage{multirow}
\usepackage{tablefootnote}
\usepackage{color}
\usepackage{algorithm}
\usepackage{algorithmic}
\usepackage{multicol}
\usepackage{pifont}

\ifCLASSINFOpdf
\else
\fi
\begin{document}
%
\title{Incorporating Heterogeneous User Behaviors and Social Influences for Predictive Analysis}
%
%
%

\author{Haobing~Liu,
        Yanmin~Zhu,~\IEEEmembership{Senior~Member,~IEEE,}
        Chunyang~Wang,
        Jianyu~Ding,
        Jiadi~Yu,~\IEEEmembership{Member,~IEEE,}
        and~Feilong~Tang,~\IEEEmembership{Senior~Member,~IEEE}
\thanks{Haobing Liu is with the Department of Computer Science and Technology, Ocean University of China, China, and the Department of Computer Science and Engineering, Shanghai Jiao Tong University, China (e-mail: haobingliu@ouc.edu.cn). Yanmin Zhu, Chunyang Wang, Jianyu Ding, Jiadi Yu, and Feilong Tang are with the Department of Computer Science and Engineering, Shanghai Jiao Tong University, China (e-mail: yzhu@sjtu.edu.cn; wangchy@sjtu.edu.cn; dingjianyu@sjtu.edu.cn; jiadiyu@sjtu.edu.cn; tang-fl@cs.sjtu.edu.cn).}
\thanks{Yanmin Zhu is the corresponding author.}
\thanks{Manuscript received June 30, 2021; revised September 26, 2021.}}

%
%

\markboth{Journal of \LaTeX\ Class Files,~Vol.~X, No.~X, June~2021}%
{Shell \MakeLowercase{\textit{et al.}}: Bare Demo of IEEEtran.cls for IEEE Journals}
%



\IEEEtitleabstractindextext{%
\begin{abstract}
Behavior prediction based on historical behavioral data have practical real-world significance. It has been applied in recommendation, predicting academic performance, etc. With the refinement of user data description, the development of new functions, and the fusion of multiple data sources, heterogeneous behavioral data which contain multiple types of behaviors become more and more common. In this paper, we aim to incorporate heterogeneous user behaviors and social influences for behavior predictions.
To this end, this paper proposes a variant of Long-Short Term Memory (LSTM) which can consider context information while modeling a behavior sequence, a projection mechanism which can model multi-faceted relationships among different types of behaviors, and a multi-faceted attention mechanism which can dynamically find out informative periods from different facets. Many kinds of behavioral data belong to spatio-temporal data. An unsupervised way to construct a social behavior graph based on spatio-temporal data and to model social influences is proposed. Moreover, a residual learning-based decoder is designed to automatically construct multiple high-order cross features based on social behavior representation and other types of behavior representations. Qualitative and quantitative experiments on real-world datasets have demonstrated the effectiveness of this model.
\end{abstract}

}

\maketitle

\IEEEdisplaynontitleabstractindextext

%
\IEEEpeerreviewmaketitle

\section{Introduction}
Behavior prediction~\cite{WenYTPS18,chen2018predictive,WangZLLZLZ20,WenLWW0WSX21} aims to predict future behaviors based on historical behavioral data. Behavior predictions are in demand in practical applications. For example, predicting students' academic performances~\cite{xu2017progressive} is needed in a university. With the prediction task, the institution can perceive students' future academic performances. Thus, the institution can identify which students might be at risk so as to intervene in time and can teach students with their corresponding aptitude. Another well-known example is recommending products to consumers~\cite{WangZLLZLZ20}. Recommendation systems benefit consumers, providers, and platforms. 

With the rapid development of the Internet and the Internet of Things, information systems of various industries are collecting a myriad of user behavioral data. According to whether behavioral data are of the same type, behavioral data can be classified into homogeneous behavioral data and heterogeneous behavioral data~\cite{ChenYCYNL19}. Due to the refinement of user data description, the development of new functions, and the fusion of multiple data sources, heterogeneous behavioral data which contain multiple types of behaviors become more common. For example, in most universities, smart cards are used as students' IDs (for accessing facilities such as libraries) and digital wallets (for payment and recharge). Consequently, entering the library, entering the dormitory, and transaction records can be got by the institution. In most E-commerce portals, systems will record clicking, adding to cart, and purchasing behaviors of consumers. User habits can be learned from multiple types of behaviors. User habits are reflected in many facets (e.g., study habits, reading habits, and rest habits). That is to say, the relationships among multiple types of behaviors are multi-faceted. User habits are deciding factors affecting prediction results. Besides, people participate in social activities to make friends with others. Studies have shown that people may change their attitudes and behaviors in response to what they think their friends might do or think, which is known as the social influence~\cite{cialdini2004social,TangSWY09}. Social influences are key factors, too. Different parts of friends influence a user in different ways. For example, a student's dining behaviors may be influenced by her classmates, and the student's studying behaviors may be influenced by her friends whose academic performances are similar to the student.

In literature, there have been a number of studies on modeling heterogeneous behavioral data for predictive analysis. According to whether models capture sequential dependencies, studies can be classified into two categories. 
The first category mainly extracts various static features from heterogeneous behaviors or extends the Matrix Factorization (MF) algorithm to handle different types of behaviors. For instance, Wang et al.~\cite{wang2015smartgpa} found correlations between students' cumulative GPAs and various features extracted from automatic sensing behavioral data. They used a generalized linear regression model as the predictive model. Fan et al.~\cite{9139346} used graph neural networks (GNNs) to model the user-item interaction graph and the user-user social behavior graph, respectively. Then operations such as concatenations were leveraged to merge representations. However, mining sequence-related patterns of behaviors cannot be achieved via these methods. 
The second category takes into account both the heterogeneity of behaviors and the sequential nature of behaviors. For instance, Kim et al.~\cite{kim2018gritnet} utilized a bidirectional LSTM to model daily activities collected in online learning environments. Zhou et al.~\cite{zhou2018micro} trained a model with a Recurrent Neural Network (RNN) to model sequential dependencies and an attention layer to capture different effects of behaviors. Tanjim et al.~\cite{tanjim2020attentive} leveraged a Transformer layer to learn item similarities from the interacted item sequence and leveraged a convolution layer to obtain the user’s intent from her actions on a particular category. Most existing studies ignore that relationships among multiple types of behaviors are diverse and model the relationships in a coarse-grained way. These methods may learn inaccurate relationships among different types of behaviors.

In this paper, we take Predicting Academic Performance (PAP)~\cite{xu2017progressive}, Predicting the Number of Borrowed Books (PNBB)~\cite{tian2011application,wang2012application}, and Predicting the Level of Financial Difficulty (PLFD)~\cite{guan2015discovery,ye2016college} as three motivating examples of behavior prediction tasks. That is to say, there are three types of target behaviors: obtaining grades, borrowing books, and applying for financial aids. Support behaviors include studying in the library, resting in the dormitory, dining in the canteen, shopping in the store, showering in the bathroom, recharging, and social behaviors. We mainly confront the following challenges. 
First, behaviors are affected by context information (e.g., weather conditions). A common scenario is that students prefer to rest in the dormitory under storming weather. A simple way to consider context information is to concatenate it with behavior information. However, we believe that context information and behavior information are different and should be treated differently. So the challenge lies in \emph{how to consider context information while modeling behaviors}. 
Second, there are multiple kinds of relationships among multiple types of behaviors, which can be found from different facets/perspectives. Most existing studies model the relationships in one latent semantic space so they cannot model the diverse relationships well. So another challenge is \emph{how to model diverse relationships among different types of behaviors}.
Third, behaviors of different periods will have different degrees of impact on one prediction task and the change varies from user to user. For instance, study plans of students vary so different days mean differently to them. So a new challenge is \emph{how to differentiate importance degrees of different periods dynamically}. 
Fourth, only by knowing social behaviors can we capture social influences. But in some cases, there is no direct access to social behaviors. For example, it is difficult to collect social behaviors on campus, while it is very easy on social media. In such cases, constructing a social behavior graph is needed. Thus, a challenge is \emph{how to construct a social behavior graph to capture social influences}.
Fifth, the relationships between social behaviors and other types of behaviors are diverse. Different parts of friends will influence one user in different ways. Most existing studies are difficult to model the diverse relationships by methods such as concatenating user representations or combining user-user graph and user-item graph into one graph. Thus, \emph{how to model diverse relationships between social behaviors and other types of behaviors} is challenging. 

To address the above challenges, we first propose a general neural network called JMBS in our preliminary work~\cite{liu2020jointly}. More specifically, each kind of daily behavior sequence is modeled by a variant of LSTM. By adding context information in the gates of LSTM, LSTM can consider context information while modeling a daily behavior sequence. Then the latent representations of all types of behaviors are concatenated. To dynamically learn different importance degrees of different days, an attention mechanism is designed. Then a final user habit representation is got. Many kinds of behavioral data belong to spatio-temporal data, and a social behavior graph can be constructed according to the co-occurrences between any pair of users. Next, a GNN is leveraged to get a social influence representation. We design a residual learning-based decoder. The decoder automatically constructs multiple high-order cross features based on the social influence representation and the habit representation, and models feature-level interactions in a fine-grained way.
In our preliminary work, we simply concatenate the latent representations of all types of behaviors assuming that these latent representations are in the same latent semantic space. In this paper, we extend our preliminary work to solve the second challenge and propose a general neural network which incorporates \underline{H}eterogeneous \underline{U}ser \underline{B}ehaviors and \underline{S}ocial influences. We call the model HUBS. Concretely, we project each type of behavior representation to multiple latent semantic spaces. In each space, vectors that are projected to the space are added together. With multiple latent semantic spaces, the model can 
model multi-faceted relationships in a fine-grained way. Besides, the proposed attention mechanisms are leveraged parallelly in multiple latent semantic spaces.

In summary, the main contributions of this paper are as follows:
\begin{itemize}
    \item We propose a general deep neural network called HUBS which incorporates heterogeneous user behaviors and social influences for predictive analysis. 
    \item We propose a projection mechanism to model the multi-faceted relationships among multiple types of behaviors in a fine-grained way. Concretely, we leverage multiple projection functions to generate multiple latent semantic spaces, i.e., latent facets. We design a multi-faceted attention mechanism to dynamically find out informative periods from different facets. Specifically, we perform multiple attention mechanisms parallelly in multiple latent semantic spaces. 
    \item We provide a taxonomy to divide existing user behavior prediction methods into four categories. Specifically, we propose two two-value dimensions to classify methods. The dimensions are homogeneous behaviors versus heterogeneous behaviors and non-sequential behavior prediction versus sequential behavior prediction. 
    \item We conduct plenty of experiments to show the effectiveness, the extensibility, and the efficiency of our model.
\end{itemize}

\begin{figure}[t]
\centering
    \includegraphics[width= 8.5cm]{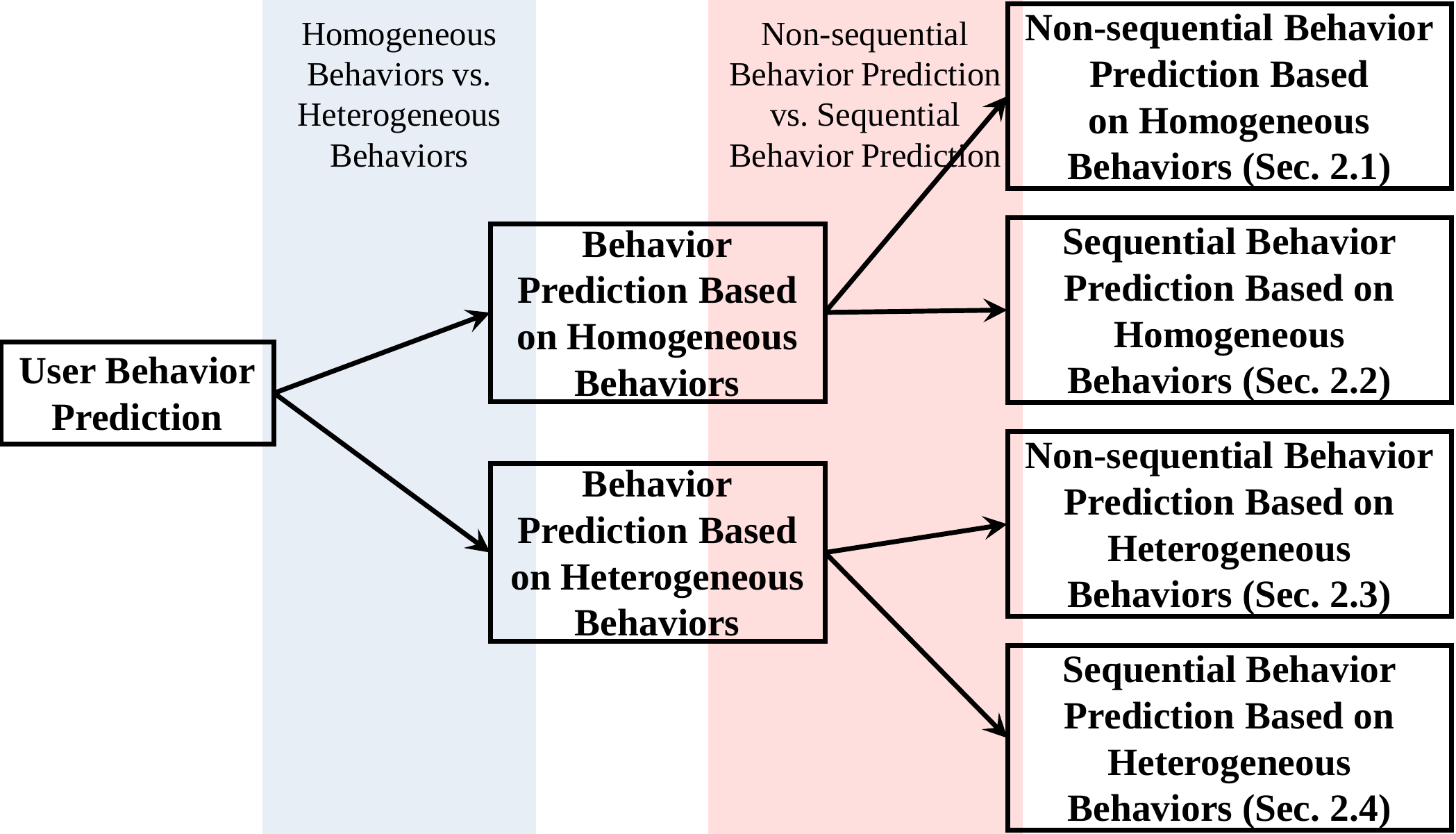}
    \caption{Classification of current studies on behavior prediction.}
    \label{fig:rw_class}
\end{figure}

The rest of this paper is organized as follows. We introduce the proposed taxonomy in Section~\ref{rw}. Next, we introduce related definitions followed by the problem statement in Section~\ref{pr}. Following that, we propose our model, i.e., HUBS in Section~\ref{mo}. Section~\ref{ex} presents the experimental results and analysis. Then, we discuss the extensibility and the efficiency of our model in Section~\ref{discu}. We discuss the ethical considerations in Section~\ref{es}. Finally, we conclude the paper in Section~\ref{co}.


\section{Related Work}\label{rw}
We provide a taxonomy to classify existing user behavior prediction methods into four categories. Specifically, we propose two two-value dimensions: \emph{homogeneous behaviors versus heterogeneous behaviors} and \emph{non-sequential behavior prediction versus sequential behavior prediction}. Figure~\ref{fig:rw_class} shows the classification.

\subsection{Non-sequential Behavior Prediction Based on Homogeneous Behaviors}
This category mainly mines the static relevancy between users and objects. According to whether leveraging deep learning, methods can be classified into traditional methods and deep learning-based methods.

\subsubsection{Traditional Methods}
Traditional Methods mainly include the MF algorithm~\cite{koren2009matrix}, the Factorization Machine (FM)~\cite{rendle2010factorization} algorithm. For example, Sweeney et al.~\cite{sweeney2015next} explored the Factorization Machine (FM)~\cite{rendle2010factorization} algorithm based on (student, course) tuple. 

Traditional methods cannot model complex nonlinear relations among features.

\subsubsection{Deep Learning-based Methods}
Some methods combine traditional methods with Multi-Layer Perceptrons (MLPs). For example, He et al.~\cite{he2017neurala} captured non-linear user-item relationships with an MLP and linear relationships with an inner product. They called the proposed architecture NCF. He and Chua~\cite{he2017neuralb} added an MLP over a FM to model high-order feature interactions and named the model NFM. Some methods are based on pooling layers. These methods regard interacted items as a set and leverage various pooling layers. For example, Zhou et al.~\cite{zhou2018deep} utilized an attention-based pooling layer to integrate items. Following the pooling layer, an MLP is adopted. Besides, some methods explore GNNs. For example, Wang et al.~\cite{wang2019neural} exploited GNNs based on user-item graphs and called the proposed model NGCF. Xu et al.~\cite{XuYSFGYZH21} constructed heterogeneous graphs whose nodes contain users, objects, and diverse attributes about objects. Some nodes might have text information. They proposed a GNN-based model to mine topic-aware semantics for learning multi-faceted node representations.

We argue that these methods model behaviors from static views and neglect the heterogeneity of behaviors.

\subsection{Sequential Behavior Prediction Based on Homogeneous Behaviors} 
This category strives to model temporal dependencies existing in behaviors. Methods can be further classified into traditional methods, Recurrent Neural Network (RNN)-based methods, Convolution Neural Network (CNN)-based methods, Transformer-based methods, GNN-based methods, and other methods.

\subsubsection{Traditional Methods}
Traditional methods mainly leverage Markov Chains (MCs). For example, Rendle et al.~\cite{rendle2010factorizing} integrated the MF with first-order MCs for next-item recommendation. Inspired by studies on knowledge graph completion, He et al.~\cite{he2017translation} regarded users as ``translation vectors'' and the previous item representation plus a user representation (i.e., a translation vector of the user) approximately determine the next item representation. 

In general, these methods can only consider short-term dependencies. 

\subsubsection{RNN-based Methods}
RNNs have been widely adopted to model sequence data and have achieved good performance in various domains (e.g., neural language processing). There are many well-known variants of RNN models, such as LSTM~\cite{hochreiter1997long}, Gated Recurrent Unit (GRU)~\cite{chung2014empirical}. Hidasi et al.~\cite{hidasi2016session} introduced GRU to recommendation systems. They leveraged a GRU to model click behaviors of users. Su et al.~\cite{su2018exercise} leveraged a bidirectional LSTM to model student exercise sequences generated in online education systems and leveraged an attention mechanism to integrate all contents. Zhu et al.~\cite{zhu2017next} proposed a variant of LSTM called Time-LSTM. Specifically, they changed the gate structure of the standard LSTM cell and the time interval is used to control the contribution of long-term memory versus short-term memory. Chen et al.~\cite{chen2018predictive} proposed another way to consider temporal information. They did not modify the internal structure of standard LSTMs. They chose to augment input representations with temporal information. 

These methods cannot model the heterogeneity of behaviors. In addition, when sequences are very long, training costs of these methods will be huge.

\subsubsection{CNN-based Methods}
CNNs have been commonly applied to process image data. Some methods leverage CNNs to handle behavioral data. For example, Tang and Wang~\cite{TangW18} regarded representations of $L$ interacted items as a $L\times d$ ``image'', where $d$ is the embedding size. Then they used a horizontal convolutional layer and a vertical convolutional layer to capture union-level patterns and point-level sequential patterns, respectively.  

CNNs are good at capturing local features. So these methods cannot capture the dependencies between two distant interactions well. 

\subsubsection{Transformer-based Methods}
Transformer~\cite{vaswani2017attention} has shown promising results in various domains. Some methods introduce Transformer to model user behaviors. For example, SASRec~\cite{kang2018self} is a self-attention network-based method and can capture pairwise item dependencies. Wang et al.~\cite{WangZLMZY21} first discovered an itemset for each interacted item with a knowledge graph. Then they merged each item representation with corresponding itemset representation. Finally, they leveraged a self-attention network to model the representation sequence.

These methods capture sequential dependencies through positional encoding technique. There exist many positional encoding schemes~\cite{qiu2021tois}. An inappropriate choice of the positional encoding scheme will lead to wrong time modeling.

\subsubsection{GNN-based Methods}
Some researchers transform sequences into directed graphs and leverage GNNs. For example, Wu et al.~\cite{WuT0WXT19} transformed sequences into directed graphs and utilized a gated graph neural network to obtain session representations. Some researchers transform sequences into undirected graphs and utilize other methods to capture sequential dependencies. For example, Fan et al.~\cite{FanLZX0Y21} mapped timestamps into vectors.

These methods rely on constructed directed graphs or other methods to capture sequential-related patterns. The transformations may be lossy since they are not one-to-one mappings.

\subsubsection{Other Methods}
There also exist some other deep learning structures that have been explored. For example, some methods combine the sliding window technique with MLPs. Wang et al.~\cite{wang2012application} leveraged an MLP to predict library circulation of next five days based on data of previous twenty days. 

\subsection{Non-sequential Behavior Prediction Based on Heterogeneous Behaviors} 
This category mainly extracts various static features from heterogeneous behaviors or extends the MF algorithm to handle different types of behaviors. Methods can be classified into traditional methods and deep learning-based methods according to whether leveraging deep learning.

\subsubsection{Traditional Methods}
Some methods first extract various static features from heterogeneous behaviors. Then they leverage traditional models. For example, Wang et al.~\cite{wang2015smartgpa} found correlations between students' cumulative GPAs and various features extracted from automatic sensing behavioral data. They used a generalized linear regression model as the predictive model. Guan et al.~\cite{guan2015discovery} identified students qualified for financial aids based on students' digital footprints. Specifically, the paper proposed a multi-label classification framework which consists of two steps: heterogeneous features extraction and funds portfolio predicting via linear model. Some methods extend classical methods to consider the heterogeneity of behaviors. For example, Zhao et al.~\cite{zhao2015improving} built matrices of different behavior types (e.g., creating post, commenting behaviors) and then they mapped every user of each behavior type and every item into one latent embedding space. Qiu et al.~\cite{qiu2018bprh} addressed multi-behavior recommendation with bayesian learning. They proposed an adaptive sampling method for the Bayesian Personalized Ranking (BPR)~\cite{rendle2009bpr} framework by considering the co-occurrence of multiple behavior types. 

These methods cannot model nonlinear interactions among features.

\subsubsection{Deep Learning-based Methods}
Some methods introduce MLPs or GNNs. For example, Sukhbaatar et al.~\cite{sukhbaatar2019artificial} employed an MLP on the set of prediction factors extracted from the online learning activities of students to identify students at risk of failing a course. Gao et al.~\cite{gao2021learning} assumed that there existed a cascading relationship among different types of behaviors and combined NCF~\cite{he2017neurala} with multi-task learning. Some researchers~\cite{jin2020multi,chen2021graph,ZhangMCX20} construct user-item multi-behavior graphs based on heterogeneous behaviors and exploit GNNs. Some researchers~\cite{LiWHH21,9139346} consider social behaviors. They design models to handle user-user graphs and user-item graphs simultaneously.  

Mining sequence-related patterns of behaviors cannot be achieved via these methods. 

\subsection{Sequential Behavior Prediction Based on Heterogeneous Behaviors} 
This category takes into account both the heterogeneity of behaviors and the sequential nature of behaviors. Studies in this category are relatively rare. Fei and Yeung~\cite{fei2015temporal} leveraged an LSTM to model student weekly activities recorded by online learning platforms. Kim et al.~\cite{kim2018gritnet} utilized a bidirectional LSTM to model daily activities collected in online learning environments. Zhou et al.~\cite{zhou2018micro} trained a model with an RNN layer to model sequential information and an attention layer to capture different effects of behaviors. Zhou et al.~\cite{zhou2018atrank} proposed a self-attention based model to learn from several behavior groups. Tanjim et al.~\cite{tanjim2020attentive} used a Transformer layer to learn item similarities from the interacted item sequence and used a convolution layer to obtain the user’s intent from her actions on a particular category. Liu et al.~\cite{liu2020learning} modeled each kind of behavior sequence with an attention-based LSTM and modeled the implicit interactions among multiple behavior prediction tasks. Liu et al.~\cite{liu2021jointly} further proposed a model to model interactions among multiple behavior prediction tasks explicitly. Wu et al.~\cite{multiviewwu22} considered a local sequence view and a global graph view. In the local view, each kind of behavior sequence was modeled by a sequence-based encoder. In the global view, a user-item multi-behavior graph was modeled by graph-based encoders. They designed three contrastive learning (CL) tasks, including the multi-behavior CL, the multi-view CL, and the behavior distinction CL. Luo et al.~\cite{LuoZYBYLQY20} constructed heterogeneous graph snapshots for each ongoing event to encode the attributes of the event and its surroundings. In addition, they combined GNNs with GRU to generate the event representation. Events are multi-typed.

Most existing methods ignore that relationships among different types of behaviors are diverse. So they model the mixed relationships directly without distinguishing the various relationships and then model them. That is to say, they model the relationships in a coarse-grained way. These methods may learn inaccurate relationships among different types of behaviors and damage the prediction performance.

\section{Preliminaries}\label{pr}
In this section, we fix some definitions and introduce the problem statement.

\noindent\textbf{Definition 1: Behavior.} While users interact with information systems, interactions will be recorded. The interactions can be semantically different. Given a target user $u$, her interactions can be defined as $X^u = \{x_1, x_2, \cdots , x_\mathcal{N}\}$. The $i$-th element $x_i = (o_i, t_i, b_i, \bm{c}_i)$ indicates that $u$ performs a behavior of type $b_i$ on the object $o_i$ at the time $t_i$. $\bm{c}_i$ is the context information which shows the environment when users perform behaviors, such as day of week. 

\begin{figure}[t]
\centering
    \includegraphics[width= 8cm]{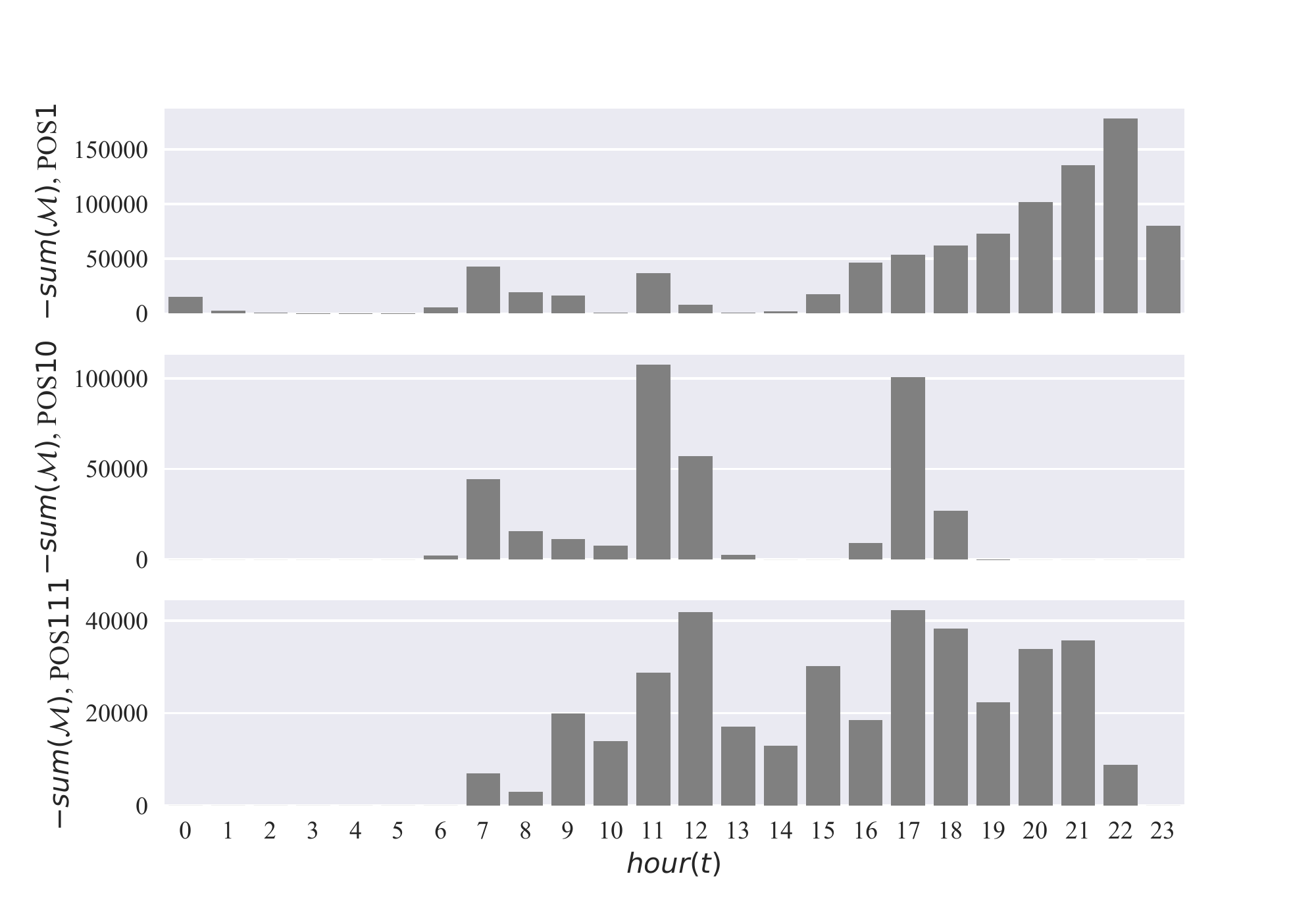}
    \caption{Distributions of students' cumulative payments at three different POS terminals with respect to the time slots.}
    \label{fig:hc}
\end{figure}
\begin{figure}[t]
\centering
    \includegraphics[width= 8cm]{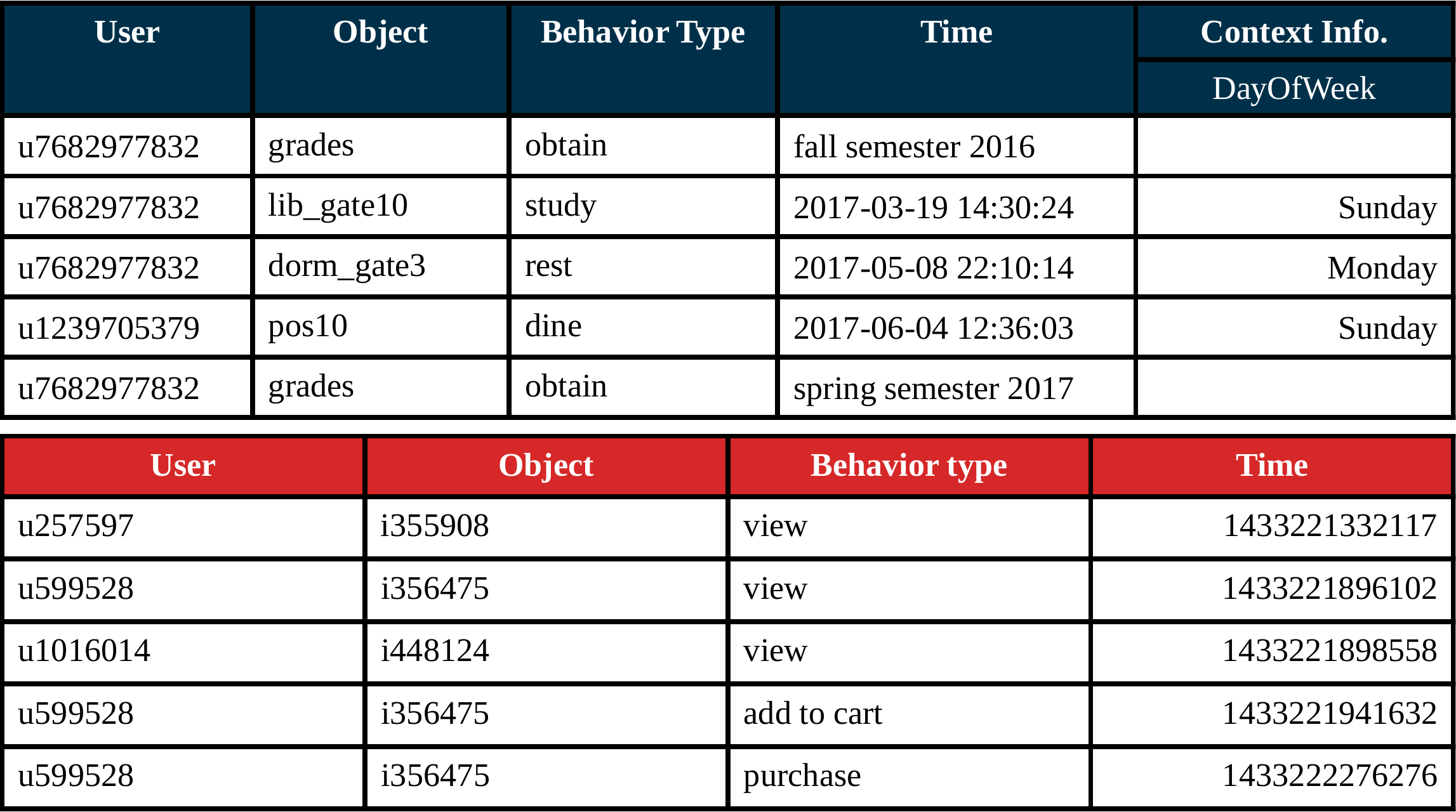}
    \caption{Heterogeneous behavioral data instances.}
    \label{fig:datexamp}
\end{figure}

It is worth noting that behavior types can be predefined depending on the actual scenario. For example, one transaction record on smart campus can be represented as $(o_i, t_i, \mathcal{M}_i)$, where $o_i$ denotes the Point of Sale (POS), $t_i$ denotes the timestamp of the record, and $\mathcal{M}_i\!\in\! \mathbb{R}^- \!\cup\! \mathbb{R}^+$ denotes the amount of money involved in the transaction record. If $\mathcal{M}_i\!<\!0$, the type of the transaction is payment; otherwise the type is recharge (i.e., \emph{recharging behavior}). We define another three types of behaviors according to the semantic functionality of $o_i$. They are \emph{dining in the canteen}, \emph{shopping in the store}, and \emph{showering in the bathroom}. To find out the semantic functionality of $o_i$, we calculate cumulative payments of students hourly for every POS terminal. Distributions at different POS terminals can be roughly classified into three categories. As Figure~\ref{fig:hc} shows, POS$1$ is set in the bathroom. Students would like to shower at night. POS$10$ is set in the canteen. Students more likely have breakfast in $[07\!:\!00,08\!:\!00)$, lunch in $[11\!:\!00,12\!:\!00)$, and dinner in $[17\!:\!00,18\!:\!00)$. POS$111$ is set in the store. Students mainly shop in $[07\!:\!00,22\!:\!00)$, extremely in $[12\!:\!00,13\!:\!00)\!\cup\![17\!:\!00,18\!:\!00)$.

For better understanding, we give examples of heterogeneous behavioral data in Figure~\ref{fig:datexamp}. There are four types of student behaviors in the upper part of the figure. In the bottom part of the figure, there are three types of consumer behaviors.

\noindent\textbf{Definition 2: Social Behavior Graph.} According to social behaviors, users are connected to form a social behavior graph, namely a social network. A social network can be defined as $G=(V,E,\bm{A})$, where $V$ is a node (i.e., student) set, $E$ is an undirected edge set which indicates the connectivity between nodes, and $\bm{A}\in\mathbb{R}^{|V|\times|V|}$ is the adjacency matrix of graph $G$.

\noindent\textbf{Definition 3: Demographic Information.} Demographic information belongs to the content information associated with users. Demographic information of user $u$ can be formalized as $\bm{D}^u$, where $\bm{D}^u$ is an attribute set about demographic information. Attributes may include province, nationality, gender, grade, and school. For simplicity, we may omit the subscripts in the following contents.

\noindent\textbf{Problem Statement.} Given heterogeneous behavioral data $X$ (Social behaviors are not included.), social behavior graph $G$, demographic information $D$, and a particular behavior type (i.e., \emph{target behavior type}), the goal is to predict next target behavior. For each behavior prediction task, there exist $M+1$ support behavior types (The $(M+1)$-th support behavior type is social behavior type.). In this paper, $M$ types of support behaviors are generated every day. Target behaviors are generated every semester. Assuming that a user (i.e., a student) has $T$ observations of the target behavior type, we leverage support behaviors generated during the first $N$ days of the $T+1$ semester and target behaviors generated during the first $T$ semesters to predict the $T+1$ target behavior ($N \! < \#\ days\ in\ one\ semester$). 

\noindent\textbf{Definition 4: Academic performance}. PAP task aims to predict academic performance. We choose Weighted Average Grade (WAG) which is on a 100-point scale to quantitatively describe the academic performance of a student in one semester. WAG can be seen as term GPA.

\noindent\textbf{Definition 5: Number of borrowed books}. PNBB task aims to predict the number of borrowed books. We count the number of borrowed books each semester.

\noindent\textbf{Definition 6: Level of financial difficulty}. PLFD task aims to predict the level of financial difficulty. The assessment on financially disadvantaged students will be done every semester. It is the prerequisite of implementing the policy of helping students with financial difficulties. In general, there are two levels of financial difficulties: high difficulty level (i.e., level 1) and medium difficulty level (i.e., level 2). These 2 levels of financial difficulties act as 2 different labels. We also introduce 1 dummy label (i.e., level 3) for indicating no financial difficulty. In fact, nearly 80\% of students do not have financial difficulty each semester.

\section{Proposed HUBS Model}\label{mo}
\begin{figure*}[t]
\centering
    \includegraphics[width= 16cm]{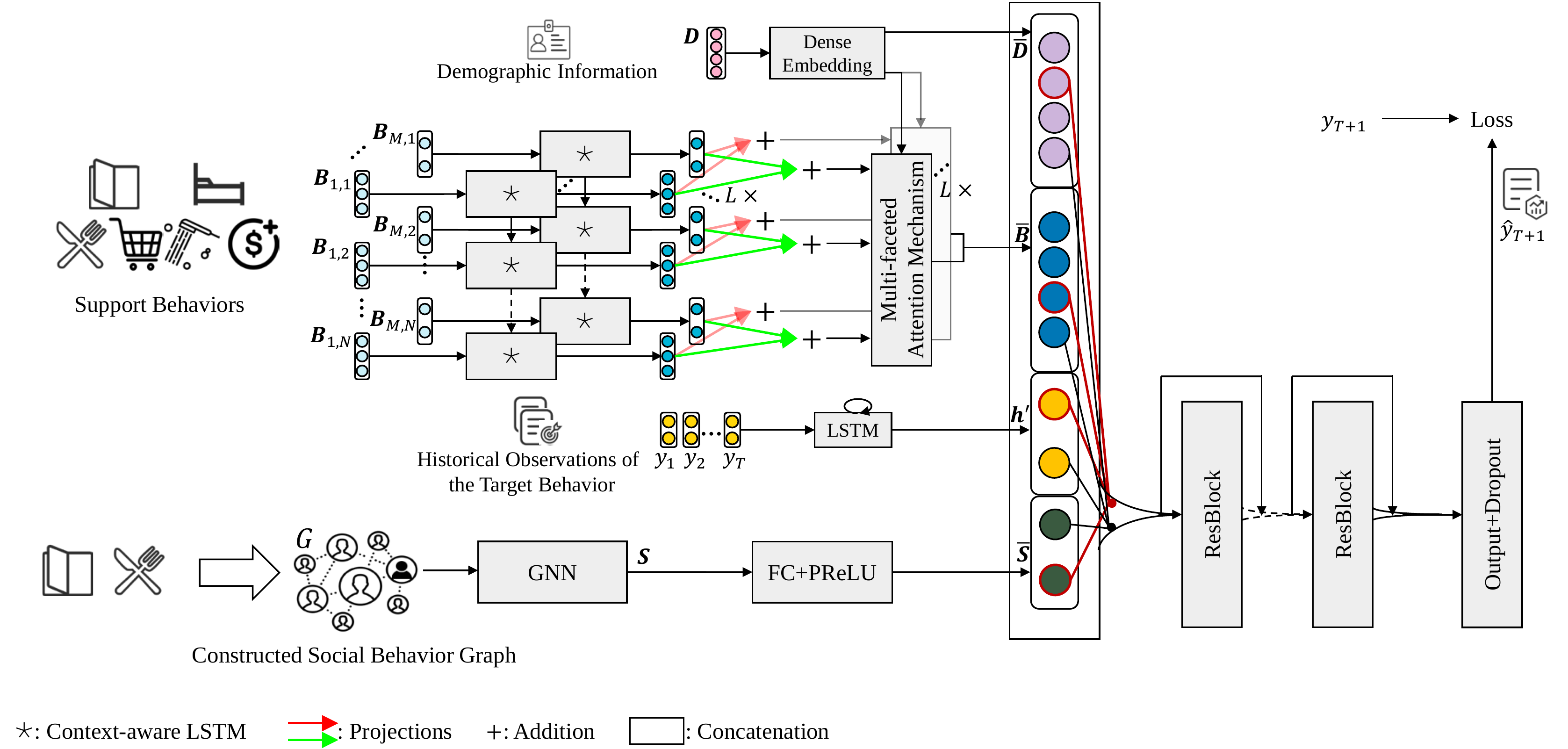}
    \caption{The architecture of HUBS. Figure best viewed in color.}
    \label{fig:structure}
\end{figure*}

The main architecture of our proposed deep neural network (HUBS) is illustrated in Figure~\ref{fig:structure}. The model consists of two parts: the user representation learning part and the interaction modeling part. Table~\ref{tab:not} summarizes main notations and their meanings used throughout this paper.

\begin{table}[t]
\scriptsize
\caption{Main notations.}
\label{tab:not}
\centering
\begin{tabular}{l|l}
\hline
Notation                  & Description                                                                                                                         \\ \hline
$\bm{D}$                  & Demographic information of the target user.                                                                                         \\
$\bm{\bar{D}}$            & User profile representation.                                                                                                  \\
$\bm{B}_{m,n}$            & Feature vector of the $m$-th support behavior type at the $n$-th day.                                                                    \\
$\bm{c}_n$                & Context information at the $n$-th day.                                                                                              \\
$\bm{h}_{m,n,\lVert_l}$   & \begin{tabular}[c]{@{}l@{}}Latent representation of the $m$-th support behavior type at the \\ $n$-th day in the $l$-th latent semantic space.\end{tabular} \\
$\bm{\alpha}_{\lVert_l}$  & Attention weights in the $l$-th latent semantic space.                                                                                     \\
$\bm{\bar{B}}_{\lVert_l}$ & User habit representation got from the $l$-th latent facet.                                                                        \\
$y_\mathcal{T}$           & Observation of the target behavior type at the $\mathcal{T}$-th semester.                                                                \\
$\bm{h}^\prime$           & Future trend representation.                                                                                                        \\
$G$                       & Constructed social network.                                                                                                         \\
$\bm{S}$                  & Node representation/social influence representation.                                                                                \\
$\bm{\bar{S}}$            & Re-trained social influence representation.                                                                                         \\ \hline
\end{tabular}
\end{table}

\subsection{User Representation Learning}
This part will learn four kinds of user representations: the user profile representation from demographic information, the user habit representation from support behaviors, the future trend representation from historical observations of the target behavior type, and the social influence representation from social behavior graph. In what follows, we introduce how to learn each kind of representation.

\subsubsection{User Profile Representation Learning}
Demographic information $\bm{D}$ is represented in the form of one high-dimensional vector: $\bm{D}=\{\bm{a}_1,\bm{a}_2,\cdots,\bm{a}_K\}$, where $K$ is the number of attributes and each attribute is represented as a one-hot vector. To reduce the dimension of $\bm{D}$, we use a dense embedding layer. The transformation is formalized as:
\begin{equation}
    \bm{\bar{D}}=[\bm{W}_{1}\bm{a}_1,\bm{W}_{2}\bm{a}_2,\cdots,\bm{W}_{K}\bm{a}_K],
    \label{d}
\end{equation}
where $\bm{W}$ terms are the mapping matrices and $[,]$ represents the concatenation of vectors. In this way, we get the user profile representation $\bm{\bar{D}}$.

\subsubsection{User Habit Representation Learning}
We divide support behaviors (Social behaviors are not included.) into several behavior groups according to behavior types. We regard one day as one time unit and aggregate behaviors within one day for each student. Let $\bm{B}_{m,n}$ denote the feature vector of the $m$-th support behavior at the $n$-th day ($1\leq m \leq M$, $1\leq n \leq N$). The way to generate $\bm{B}_{m,n}$ from $X$ is introduced as follows. We divide one day into $24$ time slots by hour (i.e., $[00\!:\!00,01\!:\!00),\cdots,[23\!:\!00,24\!:\!00)$). We find that around $100.0\%$ of studying in the library (i.e., entering the library) behaviors are generated in $[07\!:\!00,23\!:\!00)$; around $68.5\%$ of resting in the dormitory (i.e., entering the dormitory) behaviors are generated in $[17\!:\!00,24\!:\!00)\!\cup\![12\!:\!00,13\!:\!00)$. We utilize 16 elements to record entering in the library frequency in $[07\!:\!00,23\!:\!00)$ per hour at the $n$-th day; we utilize 8 elements to record entering the dormitory frequency in $[17\!:\!00,24\!:\!00)\!\cup\![12\!:\!00,13\!:\!00)$ per hour at the $n$-th day. Thus, $\bm{B}_{Lib,n}\in \mathbb{R}^{16\times 1}$ and $\bm{B}_{Dorm,n}\in \mathbb{R}^{8\times 1}$. Besides, we record the cost of each meal at the $n$-th day. Thus, $\bm{B}_{Canteen,n}\in \mathbb{R}^{3\times 1}$. To get $\bm{B}_{Store,n}$, $\bm{B}_{Bthrm,n}$ or $\bm{B}_{RP,n}$, we record the frequency and the total cost at the $n$-th day. Thus, $\bm{B}_{Store,n}\in \mathbb{R}^{2\times 1}$, $\bm{B}_{Bthrm,n}\in \mathbb{R}^{2\times 1}$, and $\bm{B}_{RP,n}\in \mathbb{R}^{2\times 1}$.

According to different tasks, proper support behaviors are chosen. Specifically, studying in the library and resting in the dormitory behaviors are chosen for PAP and PNBB tasks; dining in the canteen, shopping in the store, showering in the bathroom, and recharging behaviors are chosen for the PLFD task. Each kind of daily behavior sequence is modeled by a variant of LSTM called context-aware LSTM. Context-aware LSTM considers context information while modeling a behavior sequence and treats behavior features and context features differently at the same time. Context information $\bm{c}_{n}$ is treated as a strong signal in the gates of context-aware LSTM (as Equation~\eqref{input}, \eqref{forget} and \eqref{output} show). That is to say, context information affects what to extract, what to remember, and what to forward. 

\begin{figure}[t]
\centering
    \includegraphics[width= 6cm]{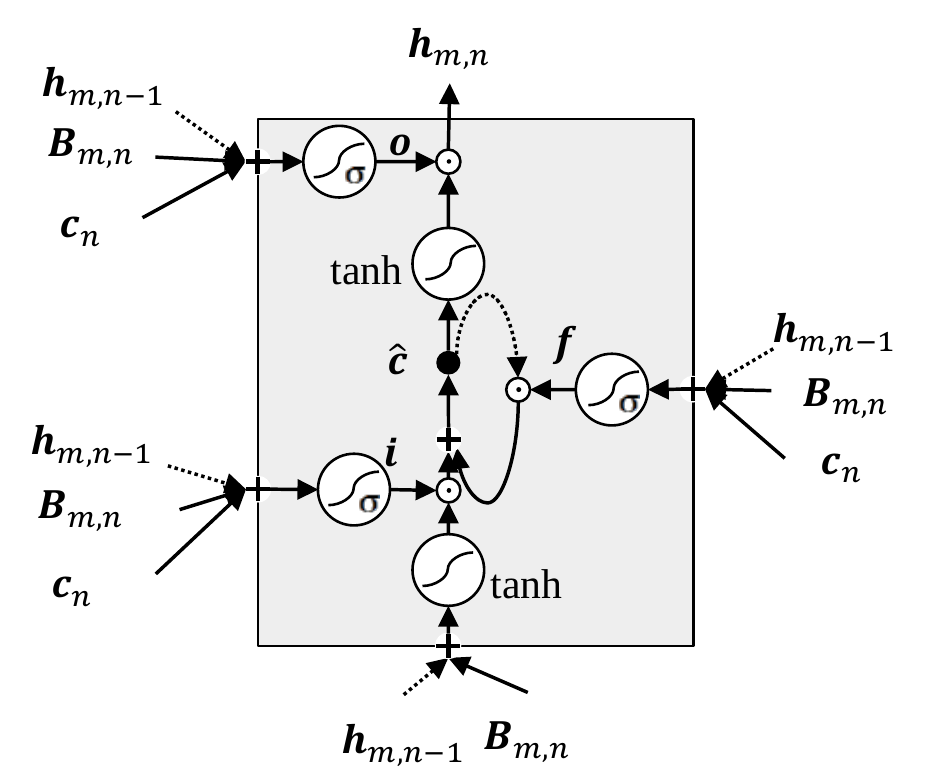}
    \caption{Detailed structure of context-aware LSTM.}
    \label{fig:clstm}
\end{figure}

Context-aware LSTM is formulated as follows and the detailed structure is shown in Figure~\ref{fig:clstm}:
\begin{footnotesize}
\begin{align}
    \bm{i}_{m,n}=&\sigma(\bm{W}_{m,iB}\bm{B}_{m,n}+\bm{W}_{m,ih}\bm{h}_{m,n-1}+\bm{W}_{m,ic}\bm{c}_{n}+\bm{b}_{m,i}), \label{input}\\
    \bm{f}_{m,n}=&\sigma(\bm{W}_{m,fB}\bm{B}_{m,n}+\bm{W}_{m,fh}\bm{h}_{m,n-1}+\bm{W}_{m,fc}\bm{c}_{n}+\bm{b}_{m,f}), \label{forget}\\
    \bm{\hat{c}}_{m,n}=&\bm{f}_{m,n}\odot\bm{\hat{c}}_{m,n-1}+\bm{i}_{m,n}\odot \tanh(\bm{W}_{m,\hat{c}B}\bm{B}_{m,n}+\notag\\ 
    &\bm{W}_{m,\hat{c}h}\bm{h}_{m,n-1}+\bm{b}_{m,\hat{c}}), \label{onlyinput}\\
    \bm{o}_{m,n}=&\sigma(\bm{W}_{m,oB}\bm{B}_{m,n}+\bm{W}_{m,oh}\bm{h}_{m,n-1}+\bm{W}_{m,oc}\bm{c}_n+\bm{b}_{m,o}), \label{output}\\
    \bm{h}_{m,n}=&\bm{o}_{m,n}\odot \tanh(\bm{\hat{c}}_{m,n}),\label{outputh}
\end{align}
\end{footnotesize}
where $\bm{i}_{m,n}$, $\bm{f}_{m,n}$ and $\bm{o}_{m,n}$ are the input, forget and output gates at the $n$-th step respectively. $\bm{\hat{c}}_{m,n}$ is the cell memory. $\bm{B}_{m,n}$ and $\bm{h}_{m,n}$ are one input and the corresponding output (i.e., hidden state) at the $n$-th step respectively. $\bm{c}_{n}$ is the context vector at the $n$-th step. $\bm{W}$ terms denote weight matrices and $\bm{b}$ terms are bias vectors. $\sigma(\cdot)$ is the element-wise sigmoid function and $\odot$ is the element-wise product.

Relationships among multiple types of behaviors (i.e., user habits) are multi-faceted. To model the multi-faceted relationships in a fine-grained way, we propose a projection mechanism. Specifically, we project each type of behavior representation to $L$ latent semantic spaces via projection functions. In each space, connections or comparisons can be made. Then one type of dependency can be automatically modeled in each space. The way of modeling relationships in multiple spaces is fine-grained. For simplicity, we leverage linear projections and the projection of the $m$-th support behavior type for the $l$-th latent semantic space can be formalized as:

\begin{equation}
    \bm{h}_{m,n,\lVert_l}=\bm{W}_{m,l} \bm{h}_{m,n}, \quad 1\leq n\leq N,
    \label{hd}
\end{equation}
where $\bm{W}_{m,l}\in\mathbb{R}^{d\times d_m}$ is the projection matrix, $\bm{h}_{m,n}\in\mathbb{R}^{d_m\times 1}$, and $\bm{h}_{m,n,\lVert_l}\in\mathbb{R}^{d\times 1}$. After the projection process, in the $l$-th latent semantic space, we add the projected vectors together:
\begin{equation}
    \bm{\bar{h}}_{n,\lVert_l}=\sum_{m=1}^M \bm{h}_{m,n,\lVert_l}, \quad 1\leq n\leq N,
    \label{hd+}
\end{equation}
where $\bm{\bar{h}}_{n,\lVert_l}\in\mathbb{R}^{d\times 1}$.

\begin{figure}[t]
\centering
    \includegraphics[width= 6cm]{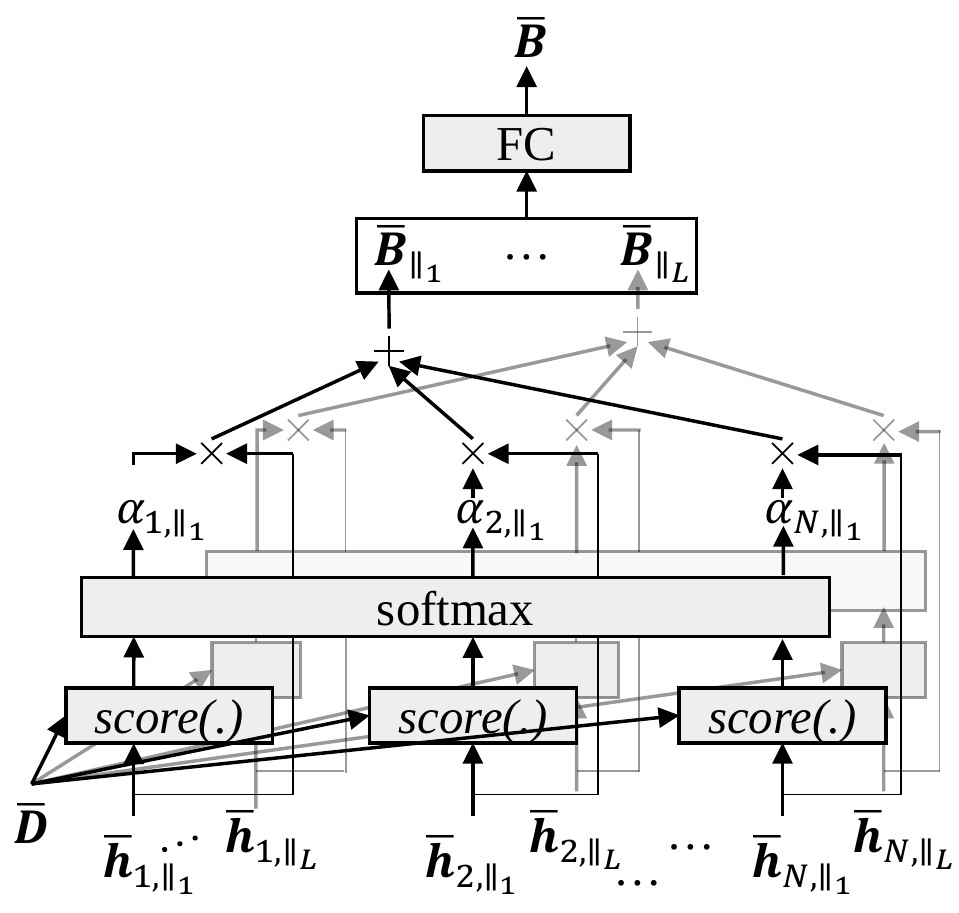}
    \caption{Detailed structure of multi-faceted attention mechanism.}
    \label{fig:att}
\end{figure}

Behaviors of different periods will have different degrees of impact in each latent semantic space. Inspired by the multi-head attention mechanism~\cite{lin2017structured,vaswani2017attention}, we design a multi-faceted attention mechanism. Specifically, in each space, we perform an attention mechanism to draw information from the sequence by different weights. Namely, we will learn $L$ different user habit representations from $L$ different latent facets. Then, these $L$ different representations are integrated to learn a final user habit representation. The structure of multi-faceted attention mechanism is illustrated in Figure~\ref{fig:att}. 

Concretely, the $l$-th representation learned from the $l$-th facet is computed with the following equations: 
\begin{small}
\begin{align}
    score(\bm{\bar{D}},\bm{\bar{h}}_{n,\lVert_l}) &=\bm{W}_{\lVert_l,a0}\tanh(\bm{W}_{\lVert_l,a1}\bm{\bar{D}}+\bm{W}_{\lVert_l,a2}\bm{\bar{h}}_{n,\lVert_l}+\bm{b}_{a}), \label{score}\\
    \alpha_{n,\lVert_l} &=\frac{\exp(score(\bm{\bm{\bar{D}},\bar{h}}_{n,\lVert_l}))}{\sum_{n^\prime} \exp(score(\bm{\bm{\bar{D}},\bar{h}}_{n^\prime,\lVert_l}))}, \label{alpha}\\
    \bm{\bar{B}}_{\lVert_l} &=\sum_{n=1}^N \alpha_{n,\lVert_l} \bm{\bar{h}}_{n,\lVert_l}, \label{bl}
\end{align}
\end{small}
where $\bm{\bar{D}}$ is calculated with Equation~\eqref{d}, $\bm{\bar{h}}_{n,\lVert_l}$ is calculated with Equation~\eqref{hd+}, $\bm{W}$ terms denote weight matrices, $\bm{b}_a$ is the bias vector, and $\bm{\bar{B}}_{\lVert_l}\in\mathbb{R}^{d\times 1}$. Next, a linear projection is leveraged to integrate learned multiple representations and calculate a final user habit representation using the following equation: 
\begin{equation}
    \bm{\bar{B}}=\bm{W}_{a3}[\bm{\bar{B}}_{\lVert_1},\bm{\bar{B}}_{\lVert_2},\cdots,\bm{\bar{B}}_{\lVert_L}],
    \label{b}
\end{equation}
where $\bm{W}_{a3}\in\mathbb{R}^{dL\times dL}$ is the projection matrix and $\bm{\bar{B}}\in\mathbb{R}^{dL\times 1}$.

\subsubsection{Future Trend Representation Learning}
Note that the length of the historical observation sequence may vary from user to user, so we adopt a dynamic LSTM to capture the temporal dependencies existing in the observation sequence and formulate it as:
\begin{equation}
    \bm{h}_\mathcal{T}^\prime=\mathrm{LSTM}(y_\mathcal{T},\bm{h}_{\mathcal{T}-1}^\prime), \quad 1\leq \mathcal{T}\leq T, \label{hT}
\end{equation}
where $y_\mathcal{T}$ and $\bm{h}_\mathcal{T}^\prime$ are one input and the corresponding output (i.e., hidden state) at the $\mathcal{T}$-th step. The final hidden state $\bm{h}_T^\prime$ is the future trend representation and we represent it as $\bm{h}^\prime$ for consistency.

\subsubsection{Social Influence Representation Learning}
Inspired by work~\cite{crandall2010inferring,yao2017predicting}, many kinds of behavioral data belong to spatio-temporal data and we infer the friendship between two users by the co-occurrences. With the increase of the co-occurrences frequency of two users, it becomes more evident that they are friends. Here, one \emph{co-occurrence} is defined as that two students enter the library (i.e., study in the library) at the same library gate within a short time interval (e.g., 10 minutes) or two students pay for meals (i.e., dine in the canteen) at the same POS terminal within a short time interval (e.g., 10 minutes). We do not utilize resting in the dormitory behaviors to compute co-occurrences because a pair of friends may not live in the same dormitory. Besides, our investigation demonstrates that almost 100\% of students prefer to dine rather than shop or shower with their friends. Thus, the friendship inferred based on co-occurrences in the library/canteen will be more accurate.

To prove our idea and find thresholds for removing the effect of random cases, we plot co-occurrence distributions in the real world/random cases in one semester as Figure~\ref{fig:sonetFig} illustrates. In particular, we simulate the random cases of co-occurrences in the library by randomly shuffling the object (i.e., library gate) or randomly shuffling the timestamp. For example, if the fragment of library entrance records is \{(stu1,7,2016-10-16 19:17:02),(stu2,2,2016-10-16 19:17:10),(stu3,2,2016-10-16 19:17:29)\}, we shuffle these three timestamps and may get \{(stu1,7,2016-10-16 19:17:29),(stu2,2,2016-10-16 19:17:10),(stu3,2,2016-10-16 19:17:02)\}. The comparison of co-occurrence distributions between the real world and random cases is shown in Figure~\ref{fig:sonetFig}(a). From the figure, we can see that when the co-occurrence frequency of two students is more than 30, these two students are likely to be friends. Thus, the threshold is set as $\frac{30\times N}{\#\ days\ in\ one\ semester}$. Similarly, we shuffle the object (i.e., POS terminal) and shuffle the timestamp. The comparison is shown in Figure~\ref{fig:sonetFig}(b). From the figure, we can see that when the co-occurrence frequency of two students is more than 20, these two students are likely to be friends. Thus, the threshold is set as $\frac{20\times N}{\#\ days\ in\ one\ semester}$. Then, we can define the friendship between student $u$ and student $v$ as:

\begin{figure}[t]
    \centering
    \subfigure[]{
        \includegraphics[width= 6.5cm]{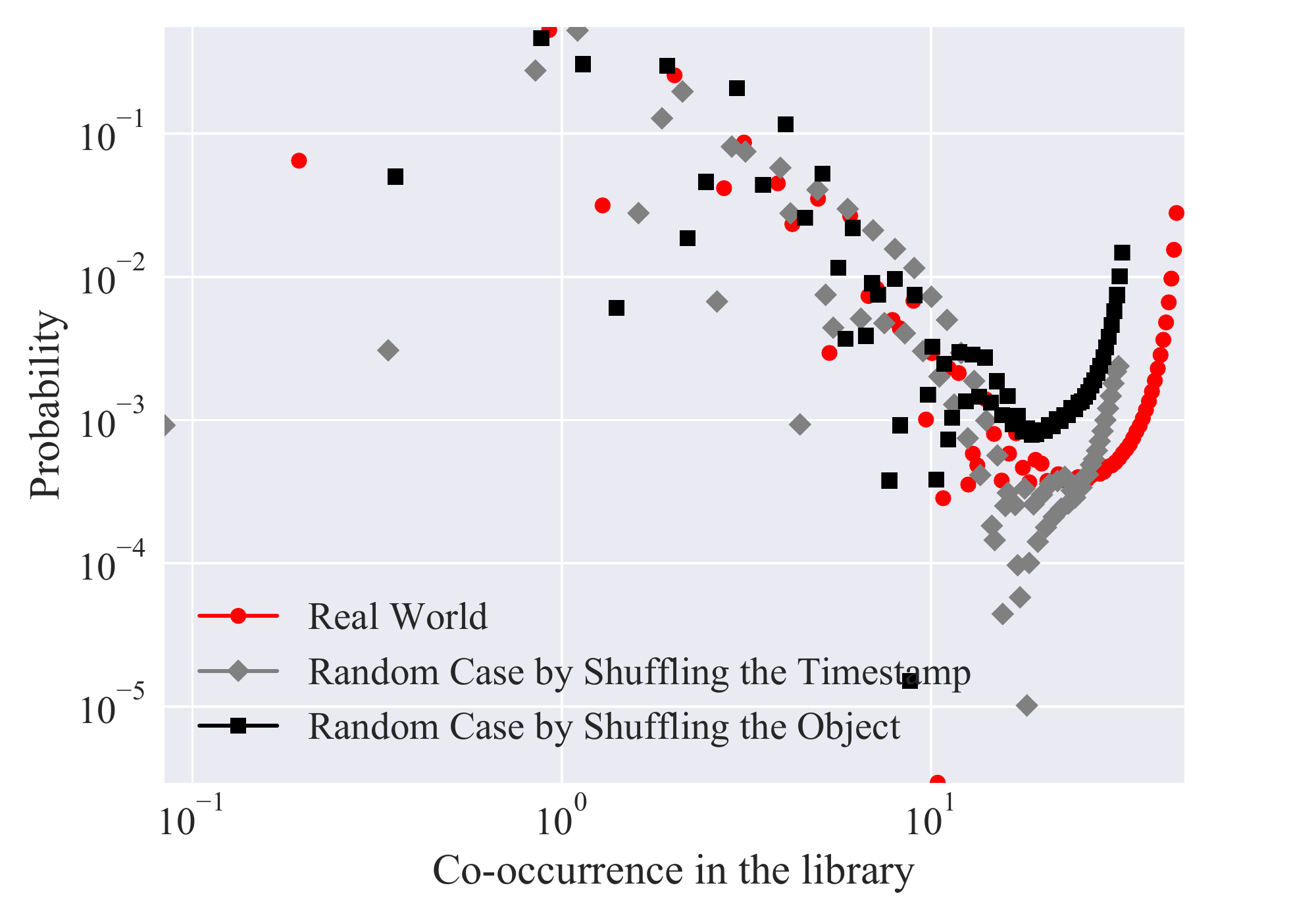}}
    \subfigure[]{
        \includegraphics[width= 6.5cm]{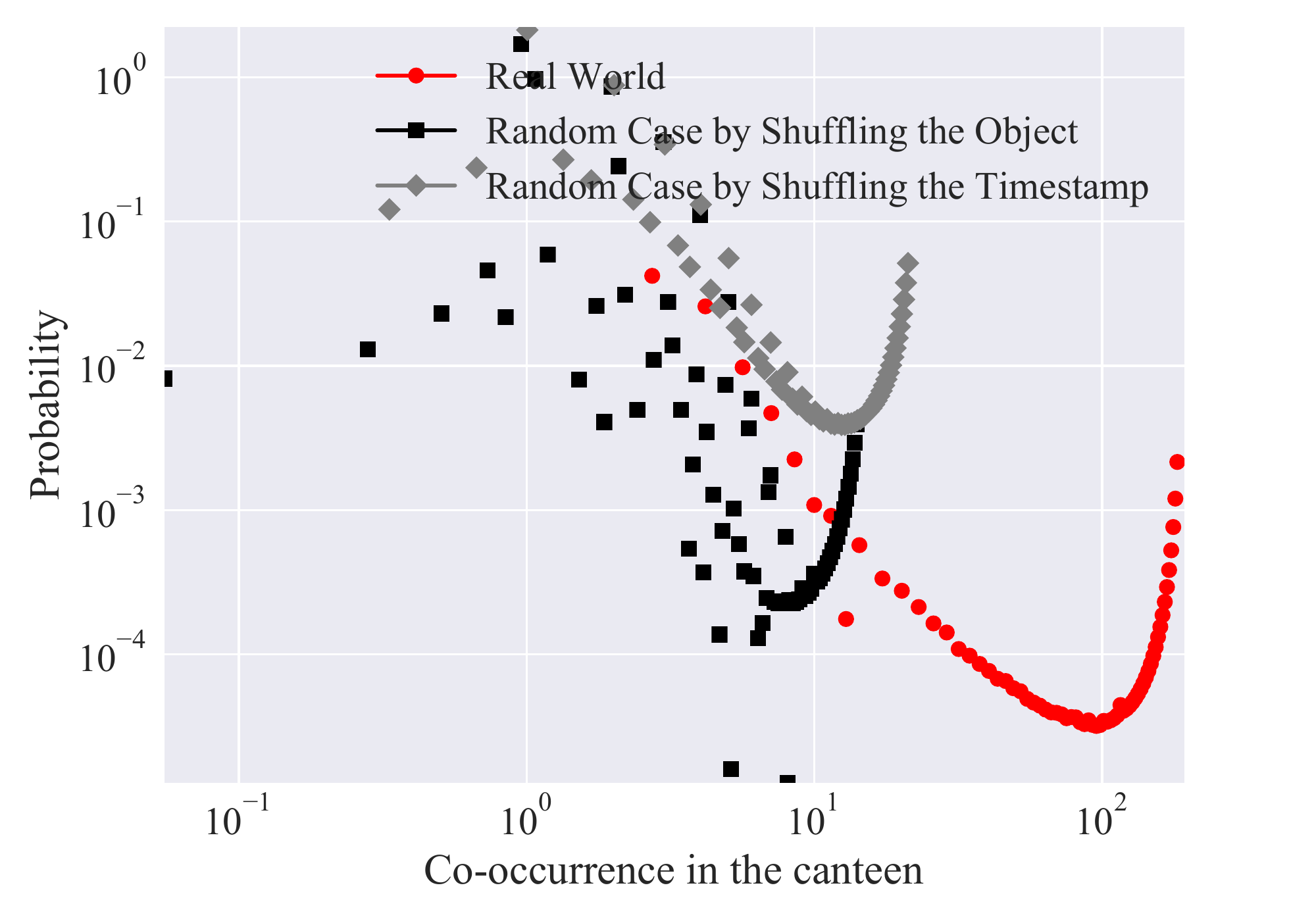}}
    \caption{co-occurrence distributions in the real world/random cases.}
    \label{fig:sonetFig}
\end{figure}

\begin{footnotesize}
\begin{equation}
friendship(u,v)=
    \begin{cases}
    1,  & \mathcal{O}_{Lib,uv}\!\ge\!\tau_{Lib} \mbox{ or }\mathcal{O}_{Canteen,uv}\!\ge\!\tau_{Canteen}, \\
    0, & \mbox{otherwise },
    \end{cases}
\label{friendship}
\end{equation}
\end{footnotesize}
\begin{scriptsize}
\begin{equation}
\tau_{Lib}=\frac{30\times N}{\#\ days\ in\ one\ semester}, \quad \tau_{Canteen}=\frac{20\times N}{\#\ days\ in\ one\ semester},\notag
\label{friendship1}
\end{equation}
\end{scriptsize}
where $\mathcal{O}_{Lib,uv}$, $\mathcal{O}_{Canteen,uv}$ are the co-occurrence frequencies of student $u$ and $v$ in the library, the canteen, respectively.

Each element in the adjacency matrix $\bm{A}$ of the social behavior graph $G$ can be calculated with Equation~\eqref{friendship}: $\bm{A}_{uv}=friendship(u,v)$.

Once getting the social behavior graph, we can apply GNNs (such as SDNE~\cite{wang2016structural}, GraphSAGE~\cite{hamilton2017inductive}, DeepInf~\cite{QiuTMDW018}) on it to embed each node into a low-dimensional vector and maintain the structural information. For simplicity, we choose to use SDNE~\cite{wang2016structural} which could capture the highly nonlinear network structure and preserve the local and global structure. For node $u$ (i.e., student $u$), SDNE outputs the embedded feature vector:
\begin{equation}
    \bm{S}^u=\mathrm{Lookup}(\mathrm{SDNE}(G),u), \label{su}
\end{equation}
where $\mathrm{Lookup}(\cdot)$ means looking up the embedded feature vector for the given identification from node embeddings.

In addition, a Fully Connected (FC) layer followed by a nonlinear activation function is employed to re-train the embedded feature vector $\bm{S}$ according to the given task:
\begin{equation}
    \bm{\bar{S}}=\mathrm{PReLU}(\bm{W}_{e}\bm{S}+\bm{b}_{e}), 
    \label{embedding}
\end{equation}
where $\bm{W}_{e}$ and $\bm{b}_{e}$ are learnable parameters, and $\mathrm{PReLU(\cdot)}$~\cite{he2015delving} is the nonlinear activation function. It is a modification of $\mathrm{ReLU}$ which replaces the zero part with a negative part controlled by learnable parameters.

\subsection{Interaction Modeling}
Relationships between social behaviors and other types of behaviors are diverse. In fact, different parts of friends influence one user in different ways. Inspire by FM-based models~\cite{rendle2010factorization,he2017neuralb} and the residual learning~\cite{he2016deep,he2016identity}, to model diverse relationships in a fine-grained way, we propose a ResBlock. According to representation learning, each dimension in a latent vector represents a feature. After getting four kinds of user representations via encoders, we feed these latent vectors into a ResBlock-based decoder. By stacking multiple ResBlocks, the decoder gradually automatically constructs multiple high-order cross features based on latent vectors. Specifically, the decoder has three types of layers: the concatenate, the ResBlock, and the output Layer.

Firstly, four kinds of user representations are concatenated into one high-dimensional feature vector as the input to the ResBlock layer:
\begin{equation}
    \bm{X}^{(0)}=[\bm{\bar{D}},\bm{\bar{B}},\bm{h}^\prime,\bm{\bar{S}}]. \label{concat}
\end{equation}

\begin{figure}[t]
\centering
    \includegraphics[width= 6cm]{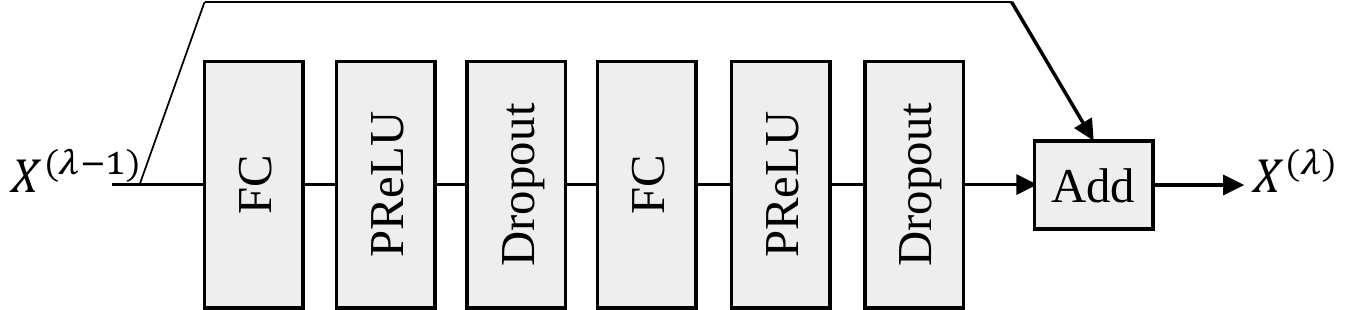}
    \caption{Detailed structure of ResBlock.}
    \label{fig:resb}
\end{figure}

Residual learning allows CNNs to have a super deep structure. Different from visual computing, our input is a high-dimensional vector and has no spatial proximity relation, thus convolution filter is not feasible. In this paper, we design a variant of residual blocks depicted in Figure~\ref{fig:resb}. Formally, one ResBlock is defined as:
\begin{equation}
    \bm{X}^{(\lambda)}=\bm{X}^{(\lambda-1)}+\mathcal{F}(\bm{X}^{(\lambda-1)}), \quad 1\leq \lambda\leq \Lambda,
    \label{resbeq}
\end{equation}
where $\mathcal{F}(\cdot)$ is the residual function (i.e., two combinations of ``FC+PReLU+Dropout'') and $\lambda$ is the block index.

The dimensions of $\bm{X}^{(\lambda-1)}$ and $\mathcal{F}(\bm{X}^{(\lambda-1)})$ must be equal in Equation~\eqref{resbeq}. If the dimensions are not equal, we can perform a linear transformation by the skip connection:
\begin{equation}
    \bm{X}^{(\lambda)}=\bm{W}_p\bm{X}^{(\lambda-1)}+\mathcal{F}(\bm{X}^{(\lambda-1)}),
    \label{resbeq2}
\end{equation}
where $\bm{W}_{p}$ is the transformation matrix.

We stack $\Lambda$ ResBlocks. Thus the decoder can automatically construct multiple high-order cross features based on social behavior representation and other types of behavior representations, and models feature-level interactions in a fine-grained way.

Finally, we feed the output of the last ResBlock to the output layer to get the final predicted value:
\begin{equation}
    \hat{y}_{T+1}=\phi(\bm{W}_o\bm{X}^{(\Lambda)}+b_o), \label{totaloutput}
\end{equation}
where $\bm{W}_o$ and $b_o$ are parameters; $\phi(\cdot)$ represents the activation function. We regard PAP task and PNBB task as regression tasks and we choose $\tanh(\cdot)$ as the activation function. Meanwhile, we regard PLFD task as a classification task and we choose $\mathrm{softmax}(\cdot)$ as the activation function.

\begin{algorithm*}[!t]
	\caption{Learning Algorithm for HUBS}
	\label{trainalg}
	\small
	\begin{multicols}{2}
	\begin{algorithmic}[1] 
    \REQUIRE ~~\\ 
        Given heterogeneous behavioral data $X$ (Social behaviors are not included.), social behavior graph $G$, demographic information $D$, and the target behavior type\\
    \ENSURE ~~\\ 
        HUBS with learned parameters $\Theta$
    \STATE Generate behavior feature vectors $\bm{B}_{m,n}$ from $X$ ($1\leq m \leq M$, $1\leq n \leq N$)
    \IF {The adjacency matrix of the social behavior graph $G$ is not available}
        \STATE Infer the adjacency matrix according to Eq.~\eqref{friendship}
    \ENDIF
    \STATE Initialize all trainable parameters $\Theta$
    \WHILE{stopping criteria is not met}
        \STATE Draw a mini-batch from the training set
        \FOR{each sample $i$}
            \STATE \emph{//User Profile Representation Learning}
            \STATE $\bm{\bar{D}}=$ Dense\_Embedding($\bm{D}$) (see Eq.~\eqref{d})
            \STATE \emph{//User Habit Representation Learning}
            \FOR{each $m \in [1,M]$}
                \STATE $\bm{h}_{m,n}=$ Context-aware\_LSTM($\bm{B}_{m,n},\bm{c}_n$), $1\leq n\leq N$ (see Eq.~\eqref{outputh})\
            \ENDFOR
            \FOR{each $m \in [1,M]$}
                \FOR{each $l \in [1,L]$}
                    \STATE $\bm{h}_{m,n,\lVert_l}=\bm{W}_{m,l} \bm{h}_{m,n}, \quad 1\leq n\leq N$ (see Eq.~\eqref{hd})\
                \ENDFOR
            \ENDFOR
            \FOR{each $l \in [1,L]$}
                \STATE $\bm{\bar{h}}_{n,\lVert_l}=\sum_{m=1}^M \bm{h}_{m,n,\lVert_l}, \quad 1\leq n\leq N$ (see Eq.~\eqref{hd+})\
            \ENDFOR
            \FOR{each $l \in [1,L]$}
                \STATE $\alpha_{n,\lVert_l}=$ Attention\_Mechanism($\bm{\bar{D}},\bm{\bar{h}}_{n,\lVert_l}$), $\quad 1\leq n\leq N$ (see Eq.~\eqref{alpha})\
                \STATE $\bm{\bar{B}}_{\lVert_l}=\sum_{n=1}^N \alpha_{n,\lVert_l} \bm{\bar{h}}_{n,\lVert_l}$ (see Eq.~\eqref{bl})
            \ENDFOR
            \STATE $\bm{\bar{B}}=\bm{W}_{a3}[\bm{\bar{B}}_{\lVert_1},\bm{\bar{B}}_{\lVert_2},\cdots,\bm{\bar{B}}_{\lVert_L}]$ (see Eq.~\eqref{b})
            \STATE \emph{//Future Trend Representation Learning}
            \STATE $\bm{h}_\mathcal{T}^\prime=\mathrm{LSTM}(y_\mathcal{T},\bm{h}_{\mathcal{T}-1}^\prime), \quad 1\leq \mathcal{T}\leq T$ (see Eq.~\eqref{hT})
            \STATE $\bm{h}^\prime\leftarrow \bm{h}_T^\prime$
            \STATE \emph{//Social Influence Representation Learning}
            \STATE $\bm{S}\leftarrow\bm{S}^i=\mathrm{Lookup}(\mathrm{SDNE}(G),i)$ (see Eq.~\eqref{su})
            \STATE $\bm{\bar{S}}=\mathrm{PReLU}(\bm{W}_{e}\bm{S}+\bm{b}_{e})$ (see Eq.~\eqref{embedding})
            \STATE \emph{//Interaction Modeling}
            \STATE $\bm{X}^{(0)}=[\bm{\bar{D}},\bm{\bar{B}},\bm{h}^\prime,\bm{\bar{S}}]$ (see Eq.~\eqref{concat})
            \FOR{each $\lambda \in [1,\Lambda]$}
                \STATE $\bm{X}^{(\lambda)}=\bm{X}^{(\lambda-1)}+\mathcal{F}(\bm{X}^{(\lambda-1)})$ (see Eq.~\eqref{resbeq})\
            \ENDFOR
            \STATE $\hat{y}\leftarrow\hat{y}_{T+1}=\phi(\bm{W}_o\bm{X}^{(\Lambda)}+b_o)$ (see Eq.~\eqref{totaloutput})
        \ENDFOR
        \IF{Regression task}
            \STATE Compute loss according to $\mathcal{L}(\Theta)=\frac{1}{|I|} \sum_{i\in I}(y^i-\hat{y}^i)^2$
        \ENDIF
        \IF{Classification task}
            \STATE Compute loss according to $\mathcal{L}(\Theta)=-\sum_{i\in I}\sum_{c\in C}(1-\hat{y}_{o,c}^i)^\gamma y_{o,c}^i\log(\hat{y}_{o,c}^i),\quad \gamma\ge0$
        \ENDIF
        \STATE Update $\Theta$ according to $\mathcal{L}$ and optimizer
    \ENDWHILE
    \RETURN $\Theta$
	\end{algorithmic}
	\end{multicols}
\end{algorithm*}

\subsection{Optimization}
For regression tasks, we adopt Mean Squared Error (MSE) as the loss function:
\begin{equation}
\mathcal{L}(\Theta)=\frac{1}{|I|} \sum_{i\in I}(y^i-\hat{y}^i)^2, \label{regloss}
\end{equation}
where $i$ denotes one training instance, $I$ is the training set, $y^i$ is the actual value, $\hat{y}^i$ is the predicted value, and $\Theta$ are all trainable parameters in the model. 

For the classification task, we adopt Focal Loss~\cite{lin2017focal} as the loss function to address the class imbalance problem:
\begin{equation}
\mathcal{L}(\Theta)=-\sum_{i\in I}\sum_{c\in C}(1-\hat{y}_{o,c}^i)^\gamma y_{o,c}^i\log(\hat{y}_{o,c}^i),\quad \gamma\ge0, \label{classloss}
\end{equation}
where $C$ denotes the class set, $y_{o,c}$ is the binary indicator of whether class $c$ is the correct classification for observation $o$, $\hat{y}_{o,c}$ is predicted probability observation $o$ is of class $c$, $\gamma$ is the tunable focusing parameter, $i$ denotes one training instance, $I$ is the training set, and $\Theta$ are all trainable parameters in the model. The focusing parameter $\gamma$ can smoothly adjust the rate at which easy examples are down-weighted. 

In order to improve the generalization capability of our model, we adopt Dropout~\cite{srivastava2014dropout} on the output layer and the FC layers existing in ResBlock. The idea of Dropout is to prevent co-adaptation by temporarily randomly making some neurons unreliable during training. Co-adaptation phenomenon may occur in learning large neural networks. Concretely, if all the parameters are learned together in a large neural network, some of the connections will have more predictive capability and these connections are learned more (other connections are ignored). 

We adopt the adaptive moment estimation (Adam)~\cite{kingma2014adam} as the optimizer. Adam is an optimization method that can compute adaptive learning rates for each parameter and converges faster. The training process is outlined in Algorithm~\ref{trainalg}.

\section{Experiments}\label{ex}
\subsection{Dataset}
\begin{table}[!t]
\scriptsize
\caption{Datasets statistics.}
\centering
\label{tab:stat}
    \begin{tabular}{lrr}
    \hline
    \multicolumn{1}{c}{\multirow{2}{*}{Item}}                                                                & \multicolumn{2}{c}{Value}                         \\ \cline{2-3} 
    \multicolumn{1}{c}{}                                                                                     & Dataset1                & Dataset2                \\ \hline
    \# Students                                                                                              & 10,000                  & 8,005                   \\ \cline{2-3} 
    \multirow{2}{*}{Time Span}                                                                               & 09/12/2016 - 01/15/2017 & 03/22/2016 - 06/26/2016 \\
                                                                                                             & 02/20/2017 - 06/25/2017 & 09/12/2016 - 01/15/2017 \\ \cline{2-3} 
    \begin{tabular}[c]{@{}l@{}}\# Library Entrance\\ Records\end{tabular}                                    & 867,571                 & \textbackslash{}        \\
    \begin{tabular}[c]{@{}l@{}}\# Dormitory Entrance\\ Records\end{tabular}                                  & 1,783,595               & \textbackslash{}        \\
    \begin{tabular}[c]{@{}l@{}}\# Smart Card Transaction\\ Records\end{tabular}                              & \textbackslash{}        & 6,655,352               \\
    \begin{tabular}[c]{@{}l@{}}\# Support behavior types\\ besides the social behav-\\ ior type\end{tabular} & 2                       & 4                       \\
    \# Demographic Records                                                                                   & 10,000                  & 8,005                   \\
    \begin{tabular}[c]{@{}l@{}}Average \# Involved\\ Semesters per Student\end{tabular}                      & $\sim$4.6               & $\sim$4.5               \\ \hline
    \end{tabular}
\end{table}

We collected two datasets in a university. One dataset covers 10,000 undergraduate students and contains library entrance records and dormitory entrance records in two continuous semesters (i.e., 09/12/2016 - 01/15/2017 and 02/20/2017 - 06/25/2017). We use this dataset to do PAP task and PNBB task. The other dataset covers 8,005 undergraduate students and contains smart card transaction records in two continuous semesters (i.e., 03/22/2016 - 06/26/2016 and 09/12/2016 - 01/15/2017). We use this dataset to do PLFD task. Dataset statistics are shown in Table~\ref{tab:stat}. We consider weather conditions and metadata (day of week) as context data.

\subsection{Baselines}
We compare our proposed method with the following models:
\begin{itemize}
    \item \textbf{HO}: We give the prediction result by the average value of historical observations for regression tasks and by the last observation value for the classification task.
    \item \textbf{LM}~\cite{wang2015smartgpa}: We choose bayesian ridge regression (i.e., with $L_2$ regularization) as the linear regression model and logistic regression as the linear classifier.
    \item \textbf{SVM/SVR}~\cite{tian2011application}: Support Vector Machine (SVM) is a maximum-margin classifier. Support Vector Regression (SVR) is a minimum-margin regression model.
    \item \textbf{RF}: Random Forest (RF) is an ensemble method with decision trees as base learners. It is based on the ``bagging'' idea.
    \item \textbf{GBDT}: Gradient Boosting Decision Tree (GBDT) is another kind of ensemble method using decision trees as base learners. It is based on the ``boosting'' idea.
    \item \textbf{MLP}~\cite{sukhbaatar2019artificial}: An MLP consists of multiple fully connected layers.
    \item \textbf{EERNNA}~\cite{su2018exercise}: EERNNA utilizes a bidirectional LSTM to model student exercise sequences generated in online education systems and leverages an attention mechanism to integrate all contents. We concatenate behavior feature vectors generated in one day with the related context feature vector into one feature vector and regard it as an ``exercise content''. 
    \item \textbf{ATRank}~\cite{zhou2018atrank}: ATRank models partitions behaviors into different behavior groups according to the behavior type and models the influence among different behavior types via a self-attention mechanism.
    \item \textbf{APAMT}~\cite{liu2020learning}: APAMT models each kind of behavior sequence with an attention-based LSTM and models the implicit interactions among multiple tasks. We adopt it to handle PAP and PNBB tasks simultaneously.
    \item \textbf{DAPAMT}~\cite{liu2021jointly}: DAPAMT models each kind of behavior sequence with an attention-based LSTM and models the interactions among multiple tasks explicitly based on the co-attention mechanism. We adopt it to handle PAP and PNBB tasks simultaneously.
    \item \textbf{JMBS}~\cite{liu2020jointly}: JMBS is proposed in our preliminary work. It leverages multiple context-aware LSTMs and an attention mechanism to model daily behavior sequences. Besides, it adopts SDNE to handle the constructed social behavior graph and get social influence representations.
\end{itemize}

\subsection{Evaluation Metrics}
For regression tasks, we use Mean Square Error (MSE) to evaluate our model. MSE penalizes large errors more heavily than the non-quadratic metrics. For the classification task, we use Macro-F1 to evaluate our model. Macro-F1 is insensitive to the imbalance of the classes. It treats classes as equal and one model cannot get a high Macro-F1 value by assigning the majority class label. The metrics are defined as:
\begin{equation}
    MSE =\frac{1}{|I|}\sum_{i\in I}(y^i-\hat{y}^i)^2,
\end{equation}
\begin{equation}
    MacroF1 =\frac{1}{|C|} \sum_{c\in C} F1_c=\frac{1}{|C|} \sum_{c\in C} \frac{2P_cR_c}{P_c+R_c},
\end{equation}
where $i$ denotes one testing instance, $I$ is the testing set, $y^i$ is the actual value, and $\hat{y}^i$ is the predicted value. $C$ denotes the class set. $F1_c$, $P_c$, and $R_c$ are F1-score, precision, and recall with respect to class $c$.

\subsection{Implementation Details}\label{id4tss}
The length of the historical observation sequence of the target behavior type is uncertain. So except for APAMT, DAPAMT and JMBS, we extract some descriptive statistics (minimum, maximum and mean) as features for regression tasks and leverage the last observation value as a factor for the classification task. Except for EERNNA, ATrank, APAMT, DAPAMT and JMBS, we extract features from support behaviors as Fei and Yang~\cite{fei2015temporal} suggested. The attention mechanism we used in EERNNA is designed as in work~\cite{yang2016hierarchical}. In addition, we extend EERNNA so that it can take demographic information into consideration.

We represent categorical features with one-hot encoding or multi-hot encoding (used for weather conditions). We process numerical inputs with the min-max normalization to ensure they are within a suitable range. For WAG and the number of borrowed books, because we use $\tanh$ as the activation function in the output layer, we scale them into $[-1, 1]$. In the evaluation, we re-scale the predicted values back to the normal values, compared with the groundtruth. For the other numerical inputs, we scale them into $[0, 1]$. On each dataset, around half of the data (i.e., the former semester) are used as the training set (accounting for 90\%) and the validation set (accounting for 10\%); the other half (i.e., the latter semester) are used for testing. 

The hyper-parameters of all methods are tuned on the validation set. We only present the optimal settings of our method are as follows. The dense embedding layer has 30 neurons in total. The dimensions of the hidden states in the context-aware LSTMs which handle $\{\bm{B}_{Lib,n}\}$, $\{\bm{B}_{Dorm,n}\}$, $\{\bm{B}_{Canteen,n}\}$, $\{\bm{B}_{Store,n}\}$, $\{\bm{B}_{Bthrm,n}\}$ and $\{\bm{B}_{RP,n}\}$ are set to 12, 6, 4, 4, 4, and 4, respectively. We set the number of latent semantic spaces to be 4 (i.e., $L=4$). For the PAP and PNBB tasks, the dimension size of each space is set to 16. For the PLFD task, the dimension size of each space is set to 8. The dimension of the hidden state in the dynamic LSTM which handles the historical observation sequence is set to 5. The dimensions of $\bm{S}$, $\bm{\bar{S}}$ are set to 16, 8, respectively. The decoder contains 2 ResBlocks (i.e., $\Lambda=2$) and each FC layer which exists in the ResBlock contains 100 neurons. The dropout rate is set to 0.4. The focusing parameter $\gamma$ is set to 2. We make predictions after the first half of the semester (i.e., $N\ =\ \frac{\#\ days\ in\ one\ semester}{2}\ =\ \frac{126}{2}\ =63$).

\subsection{Experimental Results}
We conduct four experiments. First, we compare our proposed model with advanced methods to show the effectiveness of our model. Second, we divide all students in the testing set into several groups to see the effectiveness results breaking down various student groups. Third, we do ablation studies to prove the effectiveness of the key components in HUBS. Fourth, we study how hyper-parameters impact the performance.

\begin{table}
    \caption{Comparison of different methods. The results with the best performance are marked in bold, and the results with the second best performance are marked in italics. $^\star$ represents significance level $p$-value $<0.05$ of comparing HUBS with the best baseline.}
    \label{tab:re}
    \centering
    \begin{tabular}{l|cc|c}
    \hline
    \multirow{2}{*}{Compared Methods} & PAP               & PNBB              & PLFD            \\
                                      & MSE               & MSE               & MacroF1         \\ \hline
    HO                                & 31.85             & 63.50             & 0.6813          \\
    LM                                & 27.45             & 49.64             & 0.8052          \\
    SVM/SVR                           & 26.71             & 49.18             & 0.7524          \\
    RF                                & 17.76             & 30.02             & 0.7765          \\
    GBDT                              & 17.98             & 29.73             & 0.7658          \\
    MLP                               & 17.57             & 28.36             & 0.8297          \\
    EERNNA                            & 15.67             & 28.12             & 0.8635          \\
    ATRank                            & 15.06             & 26.55             & 0.8970          \\
    APAMT                             & 14.99             & 25.72             & -               \\
    DAPAMT                            & 13.79             & 25.13             & -               \\
    JMBS                          & \textit{13.63}             & \textit{23.97}             & \textit{0.9286}          \\ \hline
    HUBS w/ Vanilla LSTM              & 13.54             & 24.00             & 0.9314          \\
    HUBS w/o Social Information       & 13.46             & 24.40             & 0.9204          \\
    \textbf{HUBS}                     & $\textbf{12.80}^\star$    & $\textbf{22.71}^\star$    & $\textbf{0.9652}^\star$ \\ \hline
    \end{tabular}
\end{table}

\subsubsection{Comparison with Baselines}
The execution of all the models is carried out five times by changing the random seeds, and we report the average results in Table~\ref{tab:re}. The two-tailed unpaired $t$-test is performed to detect significant differences between HUBS and the best baseline. From the table, we can see that HUBS achieves the best performance with the lowest MSE $12.80$ on the PAP task, the lowest MSE $22.71$ on the PNBB task, and the largest MacroF1 $0.9652$ on the PLFD task.

We can see that HO performs poorly, as it only utilizes historical observations. Other baselines further consider more information and therefore achieve better performance. Ensemble models (i.e., RF and GBDT) perform well except on the PLFD task. We think this is due to the class imbalance problem of the PLFD task. Usually, ensemble methods will be superior to individual learners by combining them together to reduce bias and/or variance of learners. MLP performs well because MLP could model complex relations among features with deep learning. EERNNA performs better than MLP. Since it can extract features automatically from daily behavior sequences with an attention-based bidirectional LSTM compared with MLP. 
ATRank performs better than EERNNA because it could consider both the diversity and sequential nature of behaviors. APAMT performs better than ATRank on PAP and PNBB tasks. We think that is because APAMT can implicitly model the interactions among multiple prediction tasks with multi-task learning. DAPAMT performs better than APAMT since it can explicitly model the interactions among multiple tasks based on the co-attention mechanism. JMBS performs better than DAPAMT. We think the reason is that JMBS can model the social influence and complex high-order nonlinear interactions between features.

It is worth mentioning that HUBS achieves $6.1\%$, $5.3\%$, $3.9\%$ relative improvements on PAP, PNBB, PLFD tasks, compared with JMBS. HUBS can model multi-faceted relationships among multiple types of behaviors with a projection mechanism. In addition, a multi-faceted attention mechanism is designed to dynamically find out informative periods from different facets.

\subsubsection{Case Studies} 
\begin{figure}[t]
\centering
    \begin{minipage}[t]{0.48\textwidth}
    \centering
    \includegraphics[width=7cm]{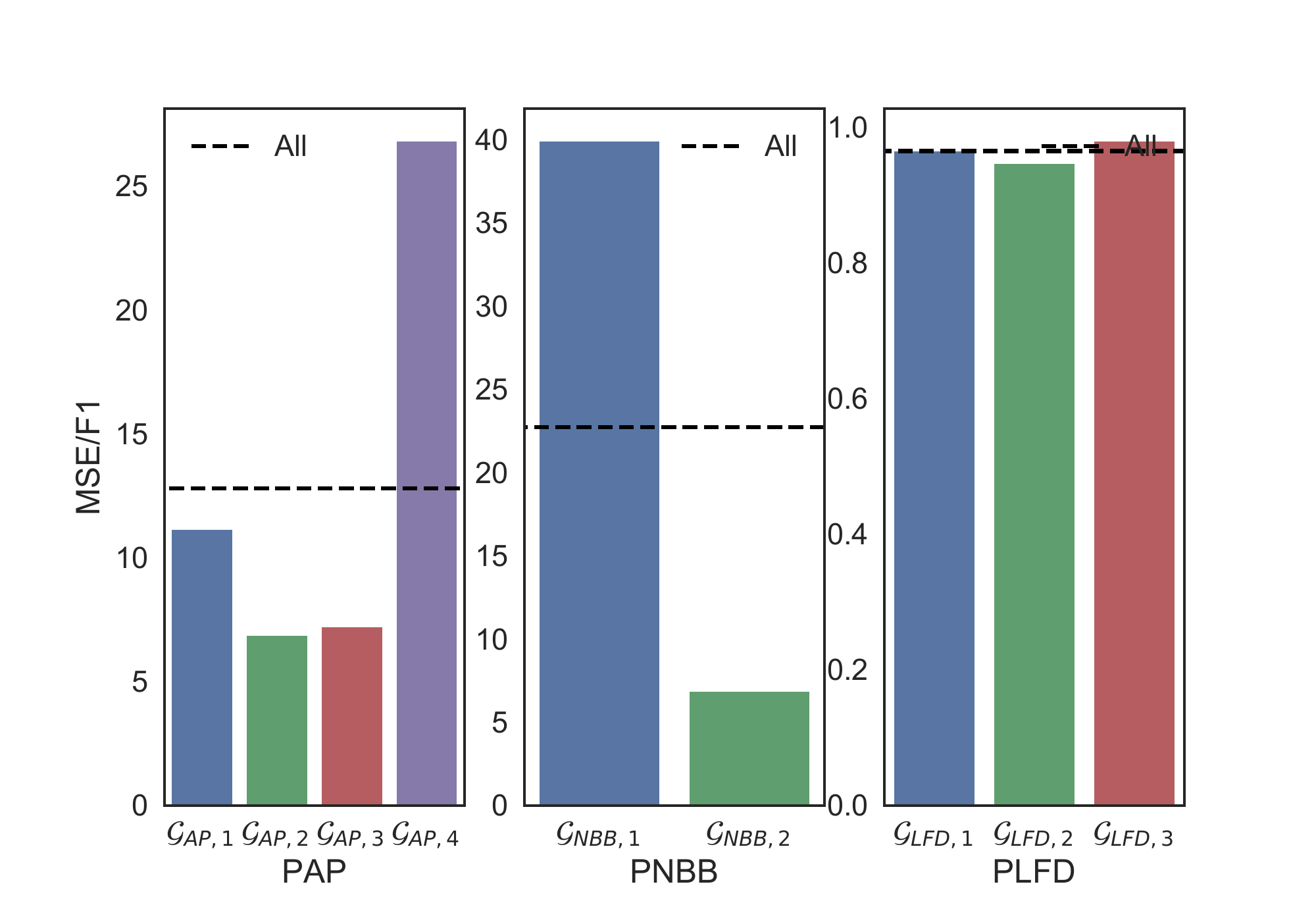}
    \caption{The effectiveness results breaking down various student groups.}
    \label{fig:casestudy}
    \end{minipage}
    \begin{minipage}[t]{0.48\textwidth}
    \centering
    \includegraphics[width=7cm]{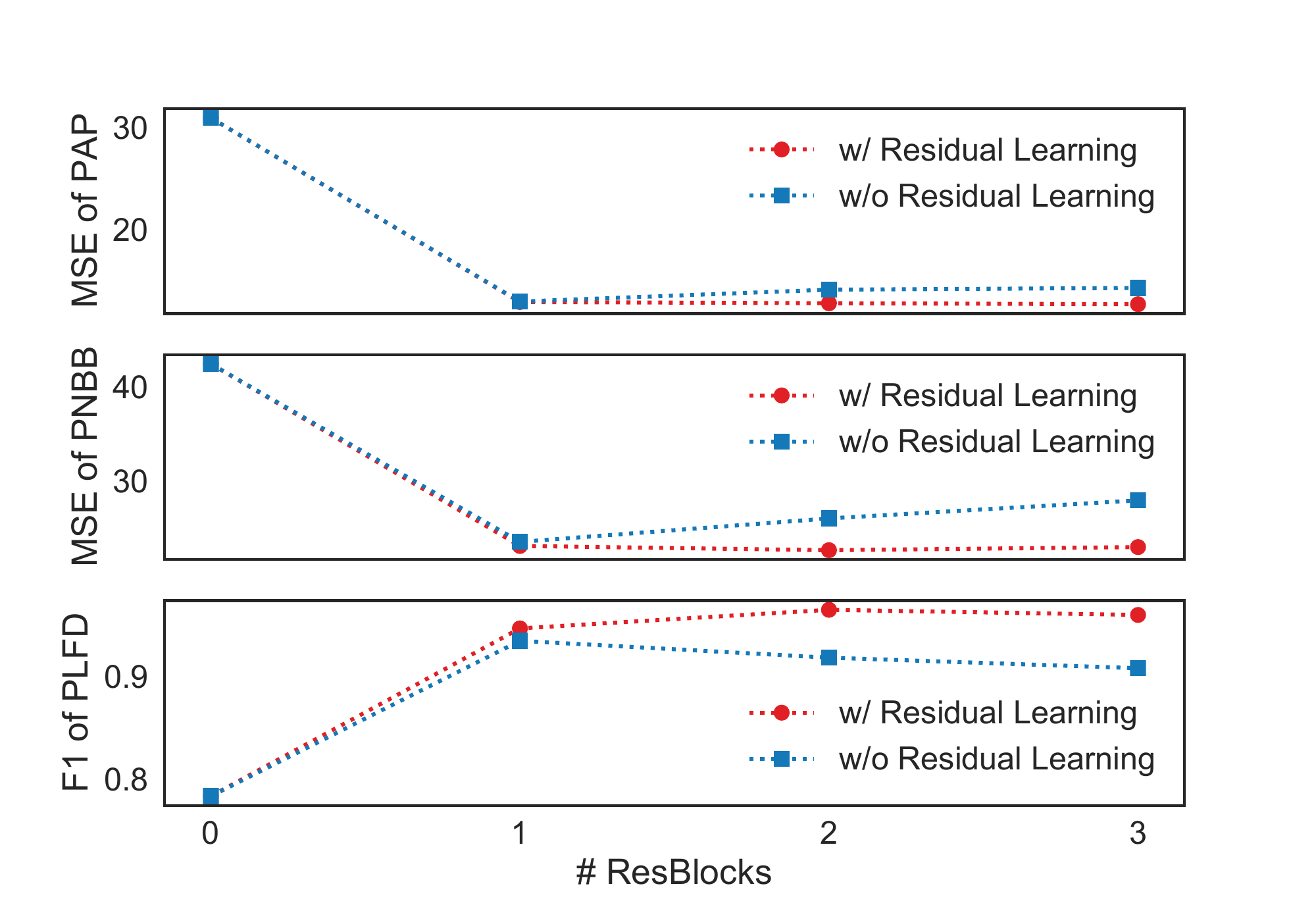}
    \caption{Performance of HUBS with respect to \# ResBlocks, whether utilizing residual learning.}
    \label{fig:res}
    \end{minipage}
\end{figure}

We divide all students in the testing set into several groups. Specifically, we rank all students by the academic performance, i.e., WAG. Then we divide all students into four groups: $\mathcal{G}_{AP,1}$ ($87.43\leq$WAG$\leq96.29$), $\mathcal{G}_{AP,2}$ ($84.28\leq$WAG$\leq87.42$), $\mathcal{G}_{AP,3}$ ($80.16\leq$WAG$\leq84.25$), and $\mathcal{G}_{AP,4}$ ($52.24\leq$WAG$\leq80.14$). Each group has the same number of students. From the left of Figure~\ref{fig:casestudy}, we can see that HUBS performs the worst on $\mathcal{G}_{AP,4}$. It is because the WAG range of $\mathcal{G}_{AP,4}$ is the biggest and the number of students who get poor academic performances is too small to train the model well. We also rank all students by the number of borrowed books. We divide students into two groups: $\mathcal{G}_{NBB,1}$ ($2\leq$\# borrowed books$\leq137$) and $\mathcal{G}_{NBB,2}$ ($0\leq$\# borrowed books$\leq1$). 48\% (around half) of students belong to $\mathcal{G}_{NBB,1}$. We can see that HUBS performs worse on $\mathcal{G}_{NBB,1}$ as the middle of Figure~\ref{fig:casestudy} illustrates. It is because the number of borrowed books range of $\mathcal{G}_{NBB,1}$ is much bigger. Finally, we divide all students into three groups according to the level of financial difficulty: $\mathcal{G}_{LFD,1}$ (level 1), $\mathcal{G}_{LFD,2}$ (level 2), and $\mathcal{G}_{LFD,3}$ (level 3). 15\% of students belong to $\mathcal{G}_{LFD,1}$ and 13\% of students belong to $\mathcal{G}_{LFD,2}$. From the right of Figure~\ref{fig:casestudy}, we can see that HUBS performs worse on $\mathcal{G}_{LFD,2}$. It is because the number of students who belong to $\mathcal{G}_{LFD,2}$ is the smallest and it is hard to train the model well.

\subsubsection{Ablation Studies} 
Table~\ref{tab:re} also provides the comparison results of variants of our proposed method. Firstly, we prove the benefit of the context-aware LSTM. We replace the context-aware LSTM with a vanilla LSTM and concatenate behavior feature vectors with related context feature vectors as inputs. The performance on the three tasks becomes poorer (reductions of $5.8\%$, $5.7\%$, $3.5\%$ on PAP, PNBB, PLFD tasks respectively). Secondly, we remove the social influence representation learning part from HUBS. The performance on the three tasks becomes poorer (reductions of $5.2\%$, $7.4\%$, $4.6\%$ on PAP, PNBB, PLFD tasks respectively). The effectiveness of our proposed residual learning-based decoder, proposed projection mechanism, and proposed multi-faceted attention mechanism is shown in Section~\ref{hps}.

\subsubsection{Effect of Hyper-parameters}\label{hps}
\noindent\textbf{Effect of the Number of ResBlocks.} As we utilize a residual learning-based decoder, problems worth studying are whether residual learning is effective and how many ResBlocks are appropriate. Experimental results are shown in Figure~\ref{fig:res}. 

As the number of ResBlocks grows, the performance grows. When the number of ResBlocks is 3, the performance on PNBB and PLFD tasks drops slightly and the performance on the PAP task barely changes. We think the overfitting problem occurs.

We remove skip connections from ResBlocks to verify the effectiveness of residual learning. Performance without residual learning is poorer than performance with residual learning and when the number of Resblocks grows, the gap becomes bigger. 

\begin{figure}[t]
\centering
    \begin{minipage}[t]{0.48\textwidth}
    \centering
    \includegraphics[width=7cm]{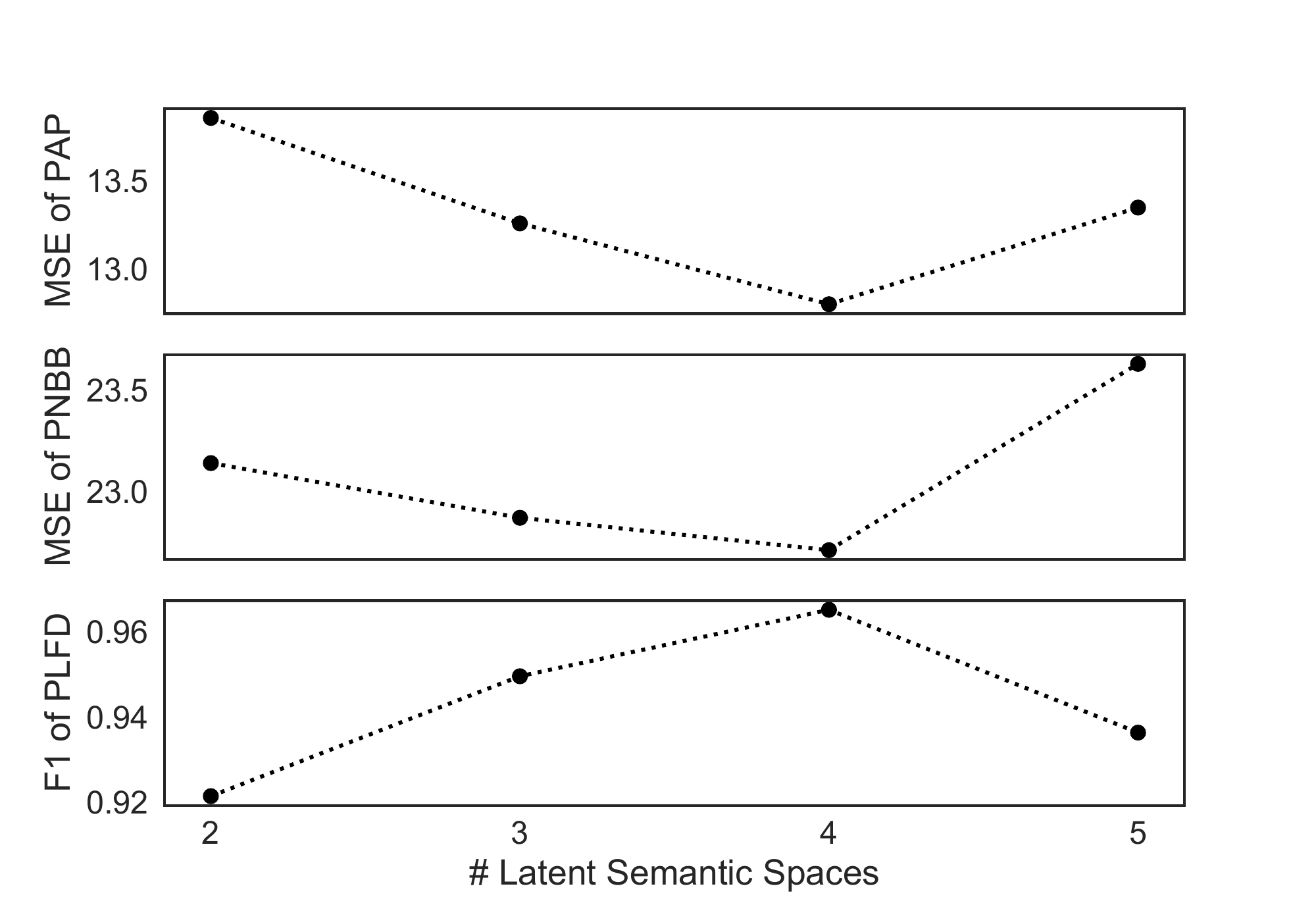}
    \caption{Performance of HUBS with respect to \# latent semantic spaces.}
    \label{fig:aspect}
    \end{minipage}
    \begin{minipage}[t]{0.48\textwidth}
    \centering
    \includegraphics[width=7cm]{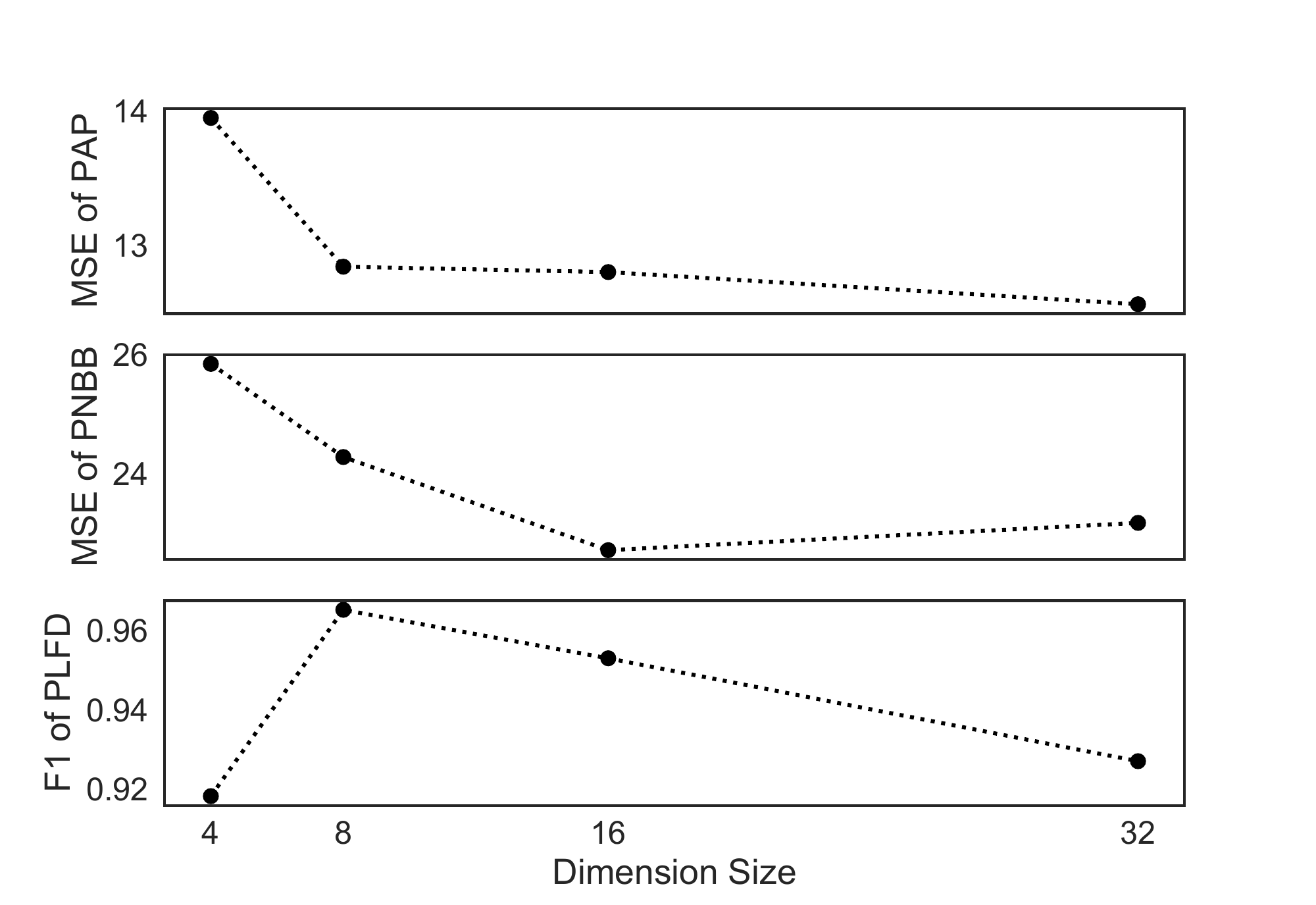}
    \caption{Performance of HUBS with respect to the dimension size of each space.}
    \label{fig:dimension}
    \end{minipage}
\end{figure}

\noindent\textbf{Effect of the Number of Latent Semantic Spaces (i.e., Facets) and the Dimension Size of Each Space.} As we project each type of behavior representation to multiple latent semantic spaces, how many spaces are appropriate and the appropriate dimension size of each space are important questions. 

First, we investigate how many latent semantic spaces are appropriate in Figure~\ref{fig:aspect}. We fix the dimension size of each space (16 for the PAP and PNBB tasks, 8 for the PLFD task) and adjust the number of spaces. As the number of spaces increases, the performance is improved. When the number of spaces is 5, the performance drops. We think the overfitting problem occurs. It is worth noting that the results also prove the effectiveness of the proposed multi-faceted attention mechanism. Thus, we set the number of latent semantic spaces to 4 in our experiments.

Next, in Figure~\ref{fig:dimension}, we study the appropriate dimension size of each latent semantic space. We fix the number of spaces as 4, and then vary the dimension size of each space. As the dimension size increases, the performance grows. When the dimension size is 32, the performance on the PNBB task drops. We think the overfitting problem occurs. At the same time, a large dimension size leads to heavy computational costs. Thus, we set the dimension size to 16 for the PAP and PNBB tasks. When the dimension size is 16, the performance on the PLFD task drops heavily. Thus, we set the dimension size to 8 for the PLFD task.

\begin{figure}[t]
\centering
    \begin{minipage}[t]{0.48\textwidth}
    \centering
    \includegraphics[width=7cm]{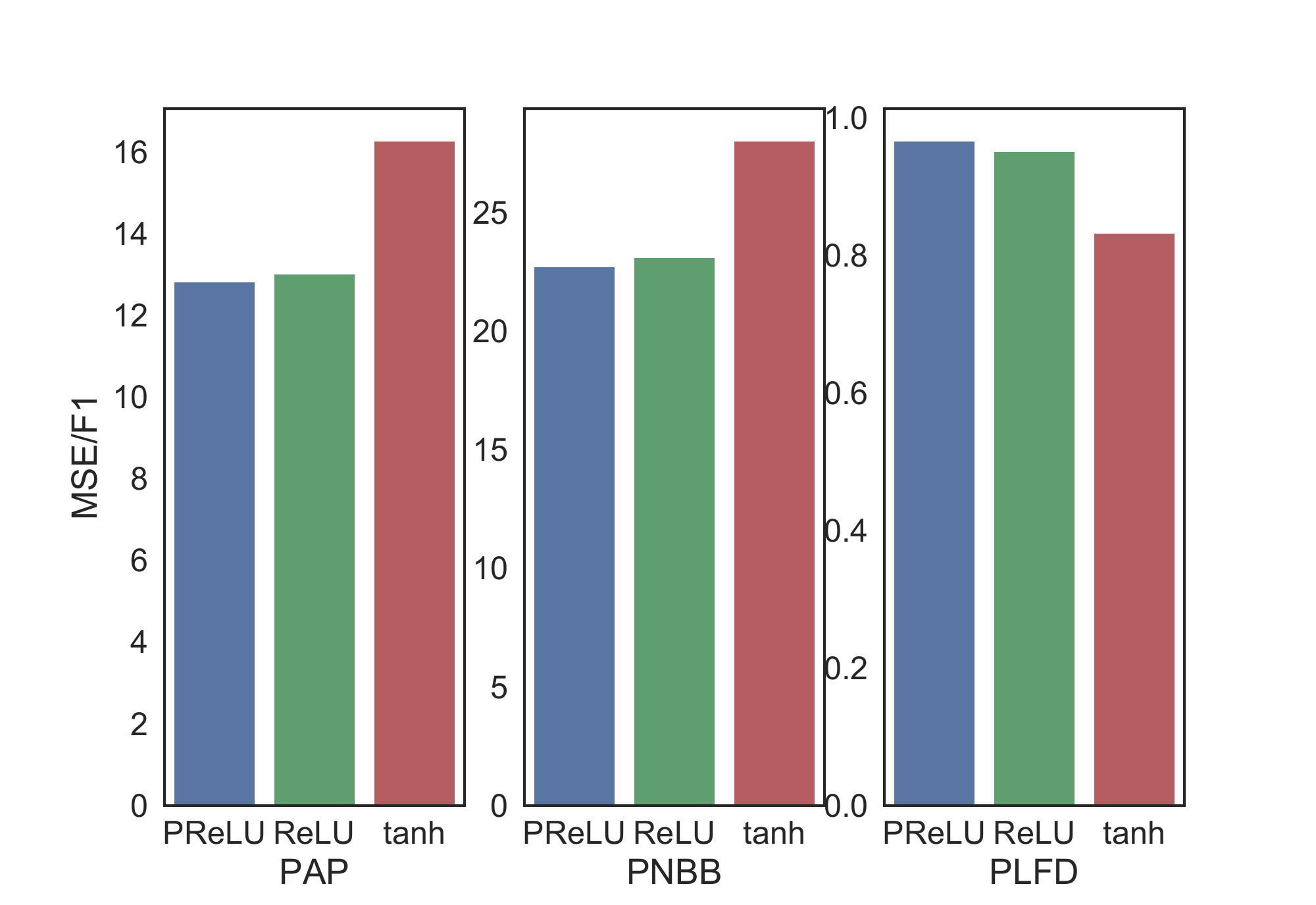}
    \caption{Performance of HUBS with respect to different activation functions.}
    \label{fig:actf}
    \end{minipage}
    \begin{minipage}[t]{0.48\textwidth}
    \centering
    \includegraphics[width=7cm]{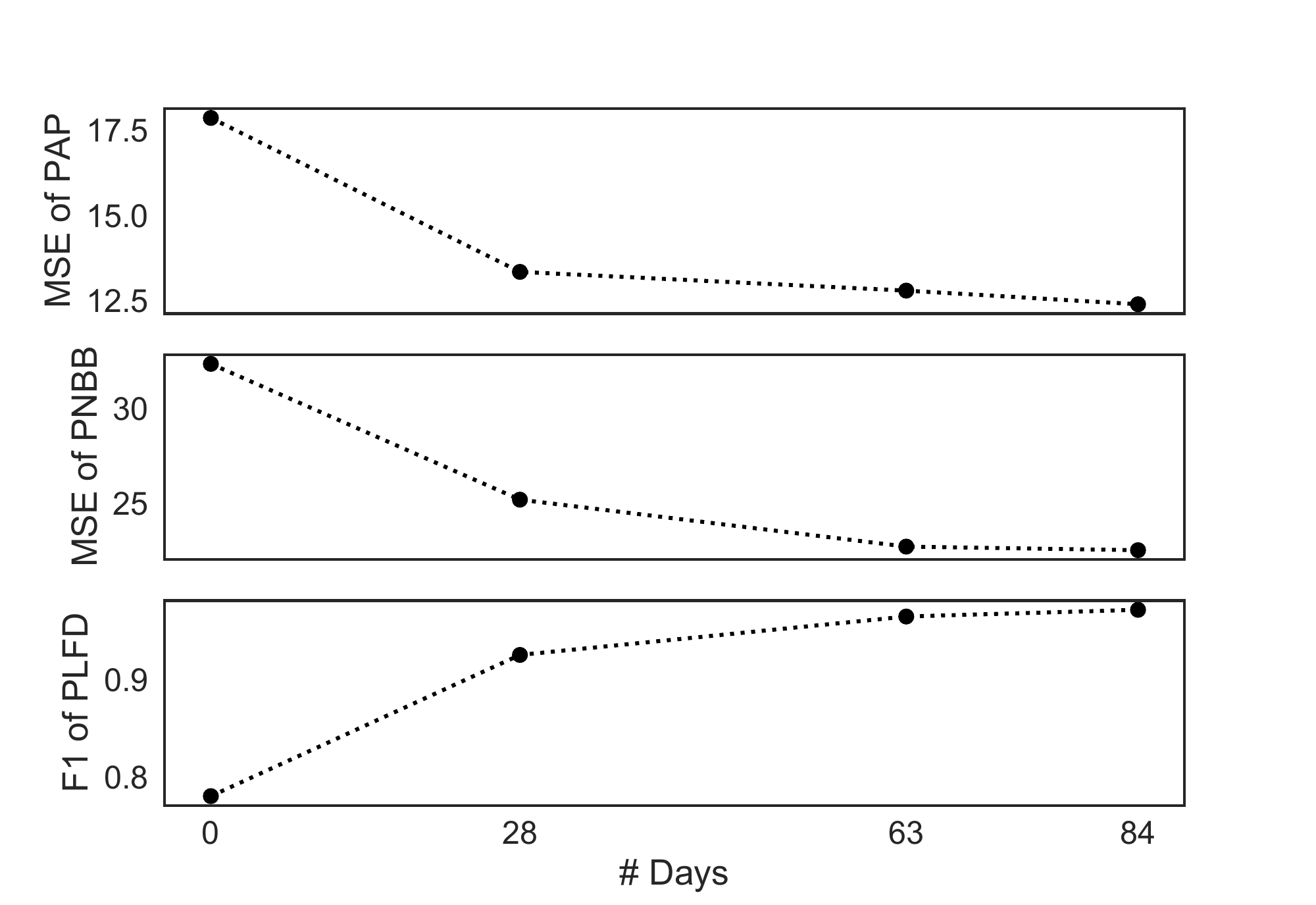}
    \caption{Performance of HUBS with respect to \# days.}
    \label{fig:days}
    \end{minipage}
\end{figure}

\noindent\textbf{Effect of Activation Functions.} We investigate the influence of the activation functions on the proposed model. We replace $\mathrm{PReLU}$ with $\mathrm{ReLU}$ or $\tanh$ in Equation~\eqref{embedding}, \eqref{resbeq}, and \eqref{resbeq2}. Figure~\ref{fig:actf} shows the results. We can see that $\mathrm{PReLU}$ is most suitable while $\tanh$ performs the worst.

\noindent\textbf{Effect of the Number of Days.} We also study the relationship between the number of days and MSE to choose the proper time for making predictions. The result is shown in Figure~\ref{fig:days}. When the number changes from 0 to 84 (i.e., 12 weeks), the performance of the model becomes better. Such a phenomenon is quite normal since a long sequence contributes to more complete information. However, the improvement is limited when the number changes from 63 (i.e., 9 weeks) to 84. Large $N$ may lead to heavy computational costs at the same time. Thus, we set $N=63$ in our experiments, thinking that making predictions after the first half of the semester is appropriate.

\section{Discussion}\label{discu}
We discuss the extensibility and the efficiency of our model in this section. In addition, we discuss the complexity for users to get started with HUBS and the effect of the training sample size.
\subsection{Model Extensibility}
We would like to note that our proposed model is a general model in the sense that it is able to model heterogeneous behaviors. $\bm{B}_{m,n}$ denotes the discrete state of the $m$-th behavior type at the $n$-th period, and the model can be generalized to model user behaviors that can be represented by discrete states. $\bm{c}_n$ denotes multi-dimension context information. Context information that is with the discrete state $\bm{B}_{m,n}$ can be included in $\bm{c}_n$. As a result, our proposed model can be extended to many other areas. 

To show the extensibility of our model, we utilize another larger dataset and we have another prediction task. The dataset, PaymentDataset, contains user payment records of 10,952 users in a community of a big company from 10/01/2016 to 06/30/2017 (i.e., three quarters) collected by a finance organization. Each record contains user identification, store identification, store category, payment time, and payment amount (USD). This dataset is utilized to predict the payment amount in a quarter ahead of time~\cite{WenYTPS18}. There are two categories of stores: restaurant and supermarket. Thus, the number of support behavior types besides the social behavior type is 2 (i.e., $M=2$). A quarter consists of around 90 days. Thus, $N=\ \frac{90}{2}\ =45$ in our experiments. There are 4,315,650 payment records for dining in the restaurant, and there are 1,155,395 payment records for shopping in the supermarket. The users' demographic information and historical quarterly payment amount are also available. We consider weather conditions and metadata (day of week) as context data. Similar to the way to generate behavior feature vectors for PLFD task, we can get $\bm{B}_{m,n}$ ($1\leq m \leq M$, $1\leq n \leq N$) based on payment records. Based on payment records for dining in the restaurant, we can infer whether two users are friends. The records of the first two quarters are used for training and the records of the last quarter are used for testing. 

We present the optimal settings of our model as follows. The total number of neurons in the dense embedding layer is 30. The dimensions of the hidden states in the context-aware LSTMs which handle dining behaviors and shopping behaviors are 4 and 4, respectively. The number of latent semantic spaces is 4 (i.e., $L=4$). The dimension size of each latent semantic space is 8. The dimension of the hidden state in the dynamic LSTM which handles the historical observation sequence is 5. The dimensions of $\bm{S}$, $\bm{\bar{S}}$ are 16 and 8, respectively. The residual network contains 2 ResBlocks (i.e., $\Lambda=2$) and each FC layer which exists in the ResBlock contains 100 neurons. The dropout rate is 0.4. 

\begin{figure}[t]
\centering
    \includegraphics[width= 5.5cm]{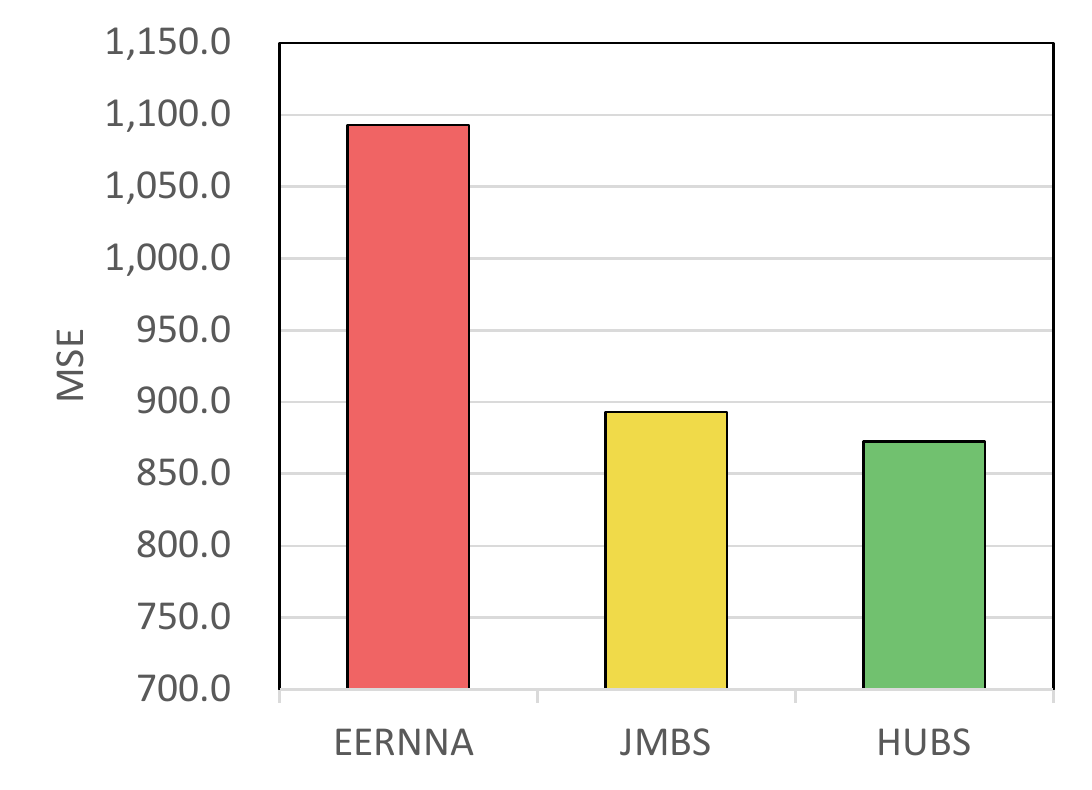}
    \caption{Comparison of competitive baselines on PaymentDataset for predicting the payment amount.}
    \label{fig:PaymentPF}
\end{figure}

We select some competitive baselines to compare. The results are shown in Figure~\ref{fig:PaymentPF}. We can see that our model still works and performs best.  

\subsection{Model Efficiency}
We then analyze the computation complexity of our model. The computational complexity of context-aware LSTMs is $O(MNd^2)$, where $d$ is the representation dimension. The computational complexity of the projection mechanism is $O(MNLd^2)$. The computational complexity of the multi-faceted attention mechanism is $O(NLd^2)$. The computational complexity of the decoder is $O(\Lambda d^2)$. Therefore, the total computation complexity of our model is $O(MNd^2+MNLd^2+NLd^2+\Lambda d^2)$. The computation complexity of JMBS is $O(MNd^2+Nd^2+\Lambda d^2)$. The computation complexity of DAPAMT is $O(MNd^2+Nd^2+\Upsilon(Zd^2+Z^2d))$, where $Z$ is the number of tasks and $\Upsilon$ is the number of Multi-task Interaction Units. In our experiments, $N>d$, $N>>M$, $N>>L$, $N>>\Lambda$, $N>>Z$, and $N>>\Upsilon$. Thus, the total computation complexity of our model, JMBS, and DAPAMT can be written as $O(MNLd^2)$, $O(MNd^2)$, and $O(MNd^2)$, respectively. $L$ is not large (less than 5 in our cases), hence the time complexity for learning HUBS is acceptable compared with other state-of-the-art models. Besides, $M$ is either not large (less than 5 in our cases). Compared with simple machine learning algorithms (e.g., MLP, GBDT), HUBS can extract features automatically from daily behavior sequences, thus the time of manual feature extraction is reduced. In fact, the manual feature extraction not only is time-consuming but also requires significant domain knowledge.

\begin{table}[t]
\caption{Running time comparison.}
\centering
\label{tab:time}
\begin{tabular}{l|cc}
\hline
\multirow{2}{*}{Methods} & PAP         & PNBB        \\
                         & Time (s)    & Time (s)    \\ \hline
APAMT                    & \multicolumn{2}{c}{12.52} \\ \cline{2-3} 
DAPAMT                   & \multicolumn{2}{c}{12.54} \\ \cline{2-3} 
JMBS                     & 12.51       & 12.51       \\
HUBS                     & 13.17       & 12.89       \\ \hline
\end{tabular}
\end{table}

We further record the running time per epoch. Experiments are run on a CPU server with 4 cores (Intel Xeon Gold 6278c @ 2.60GHz). The batch size is fixed at 32 for all methods. The results are shown in Table~\ref{tab:time}. Although the efficiency of HUBS is slightly worse than competitive baselines, the performance of HUBS is obviously better. Moreover, our model is more suitable for behavioral prediction with lower response time requirements. 

\subsection{Complexity for Users to Get Started with HUBS}
It is easy for users to get started with HUBS. Because there are not too many hyper-parameters and some hyper-parameters do not need to be tuned (These hyper-parameters keep the same setting on multiple tasks.).

Based on four behavior prediction tasks (i.e., predicting academic performance, predicting the number of borrowed books, predicting the level of financial difficulty, and predicting the payment amount), we find the number of ResBlocks should be set to 2 and each FC layer which exists in the ResBlock should contain 100 neurons. $\mathrm{PReLU}$ is the most suitable activation function in Equation~\eqref{embedding}, \eqref{resbeq}, and \eqref{resbeq2}. Making predictions after the first half of the semester/quarter is appropriate. The dimension of the social influence representation should be set to 16. Four latent semantic spaces (i.e., facets) may work well. Besides, it is worth noting that the threshold for removing the effect of random cases is not a hyper-parameter. We can know its value from data. We think users who want to tune hyper-parameters can pay attention to the dimension size of each latent semantic space. If they are not satisfied with the performance of HUBS, they may change the number of latent semantic spaces.

\subsection{Effect of the Training Sample Size}
As described in Section~\ref{id4tss}, there are $10000\times 90\%=9000$ training samples and 9000 testing samples on Dataset1; there are $8005\times 90\% \approx 7205$ training samples and 8005 testing samples on Dataset2. We select some state-of-the-art models to analyze how the training sample size affects the performance of models. We randomly select $\frac{1}{4}$ training samples, $\frac{1}{2}$ training samples, and $\frac{3}{4}$ training samples to train models.

\begin{figure}[t]
    \centering
    \subfigure[]{
        \includegraphics[width= 5.5cm]{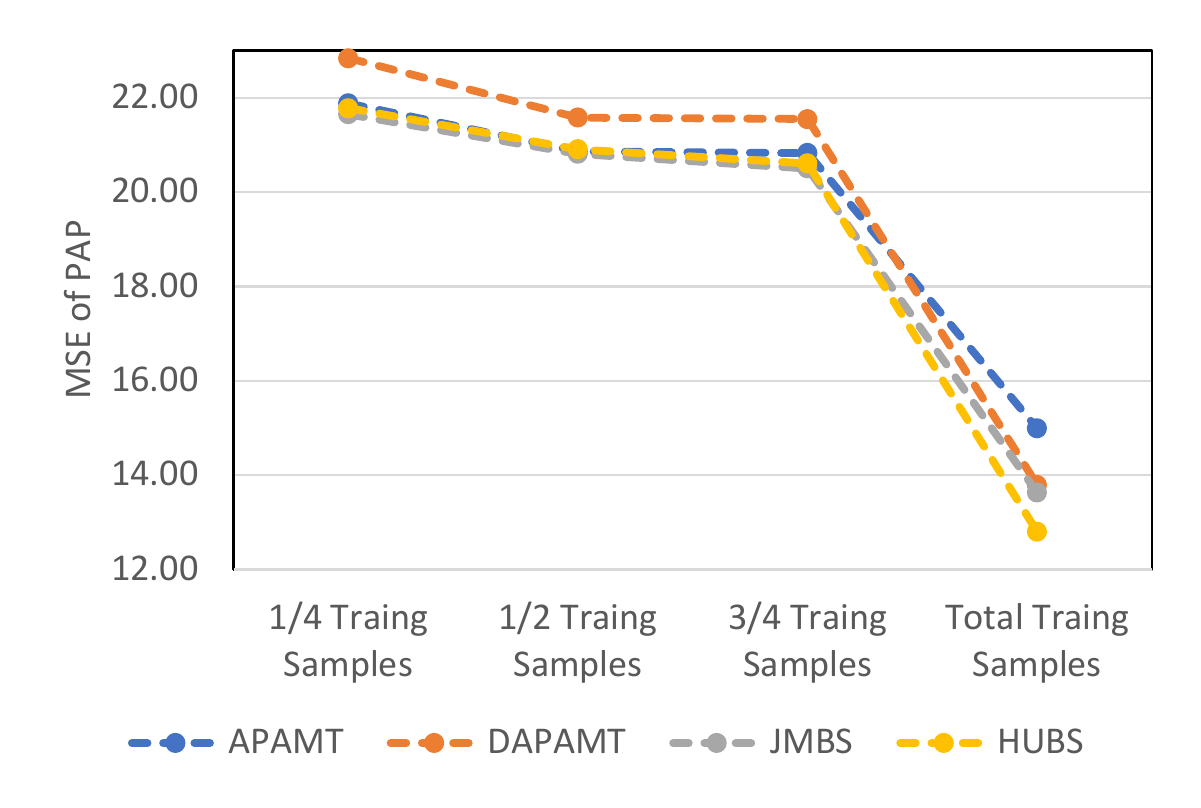}}
    \subfigure[]{
        \includegraphics[width= 5.5cm]{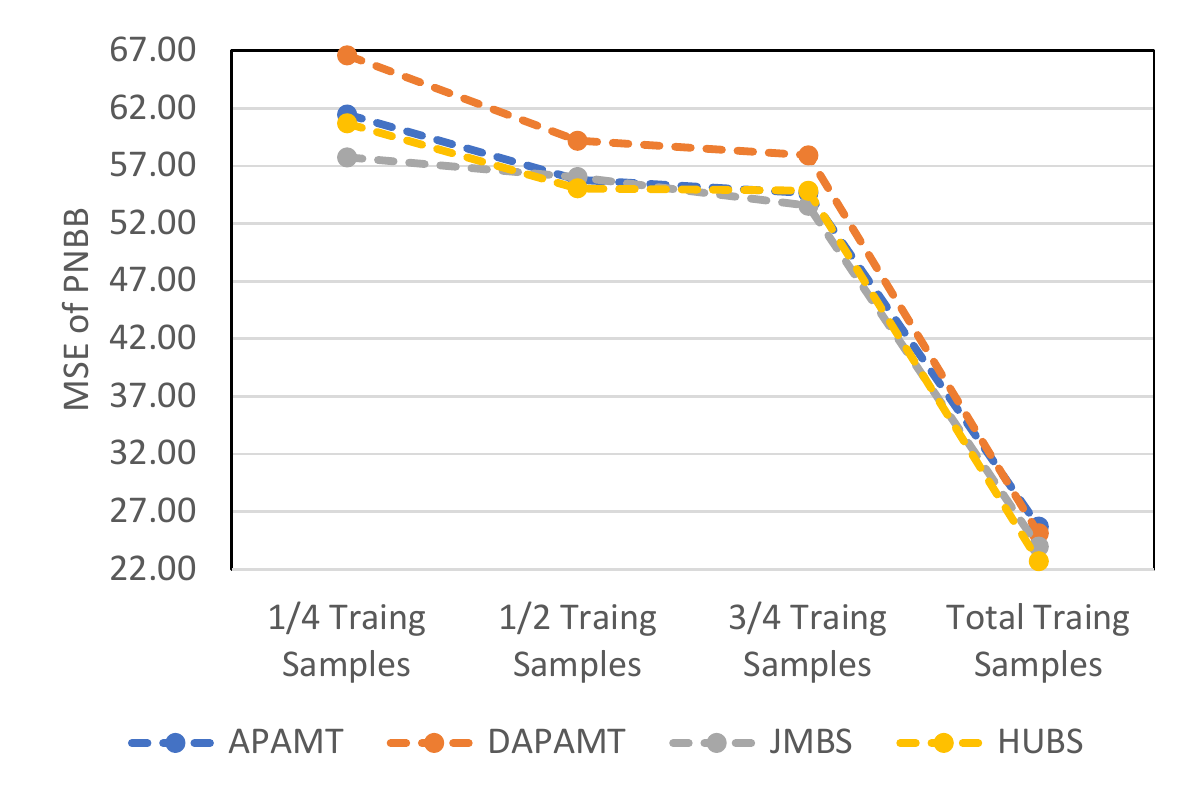}}
    \caption{Performance of models with respect to the the training sample size.}
    \label{fig:tss}
\end{figure}

Figure~\ref{fig:tss}(a) and (b) show the results on PAP and PNBB tasks, respectively. We can see that more training samples lead to better performance of models. Such a phenomenon is normal. Models are data-driven. More training samples will train models better. Besides, it is very hard to predict unseen users' behaviors (i.e., the user cold-start problem). We can also see that the performance of HUBS is a little poorer than JMBS and the performance of DAPAMT is a little poorer than APAMT when training samples are not sufficient. We think that is because HUBS and DAPAMT have more parameters to learn compared with JMBS and APAMT, respectively.

\section{Ethical Considerations}\label{es}
It is crucial to consider ethical implications. Three aspects~\cite{price2019privacy} are involved in this ethical analysis: the types of data; who will be accessing the data; and for what purpose. We explore these questions.

For the types of data, the datasets used in our work have been processed and fully anonymous. User names are removed. For those who have access to the data, the data used in our work are not released for public access. Only after privacy concerns are cleared, will the data be used for other research purposes. In applications of our prediction model, we stress that prediction results should be constrained in a small group of official staff. For the purpose of this work, we would like to stress that the use of data and predicted results should be constrained for the purpose of user management only, not for general public access. This work has been approved by the Institutional Review Board (IRB).

\section{Conclusion}\label{co}
In this paper, we propose a general deep neural network called HUBS. HUBS can jointly model heterogeneous behaviors and social influences for predictive analysis. We design a projection mechanism for modeling multi-faceted relationships among multiple types of behaviors in a fine-grained way. We design a multi-faceted attention mechanism for dynamically discovering informative periods from different facets. Experiments and analyses have demonstrated the effectiveness, extensibility, and efficiency of our model.

\section*{Acknowledgements}
We thank the anonymous reviewers for carefully reviewing and useful suggestions. This research is supported in part by the 2030 National Key AI Program of China 2018AAA0100503, National Science Foundation of China (No. 62072304, No. 61772341, No. 61832013, No. 62172277, No. 62172275), Shanghai Municipal Science and Technology Commission (No. 19510760500, No. 21511104700, No. 19511120300), the Oceanic Interdisciplinary Program of Shanghai Jiao Tong University (No. SL2020MS032), Scientific Research Fund of Second Institute of Oceanography, the open fund of State Key Laboratory of Satellite Ocean Environment Dynamics, Second Institute of Oceanography, MNR, GE China, and Zhejiang Aoxin Co. Ltd.


%



\ifCLASSOPTIONcaptionsoff
  \newpage
\fi




\bibliographystyle{unsrt}
\end{document}